\documentclass[11pt,en,bibtex,bibend=bibtex]{elegantpaper}

\usepackage{fancyhdr}
\usepackage{graphicx}
\usepackage{float}
\usepackage{amsfonts} 

\newcommand{\logoheight}{0.8cm}   
\newcommand{\logofile}{logo_text.pdf}  
\fancypagestyle{plain}{
    \fancyhf{} 
    \fancyhead[L]{\includegraphics[height=\logoheight]{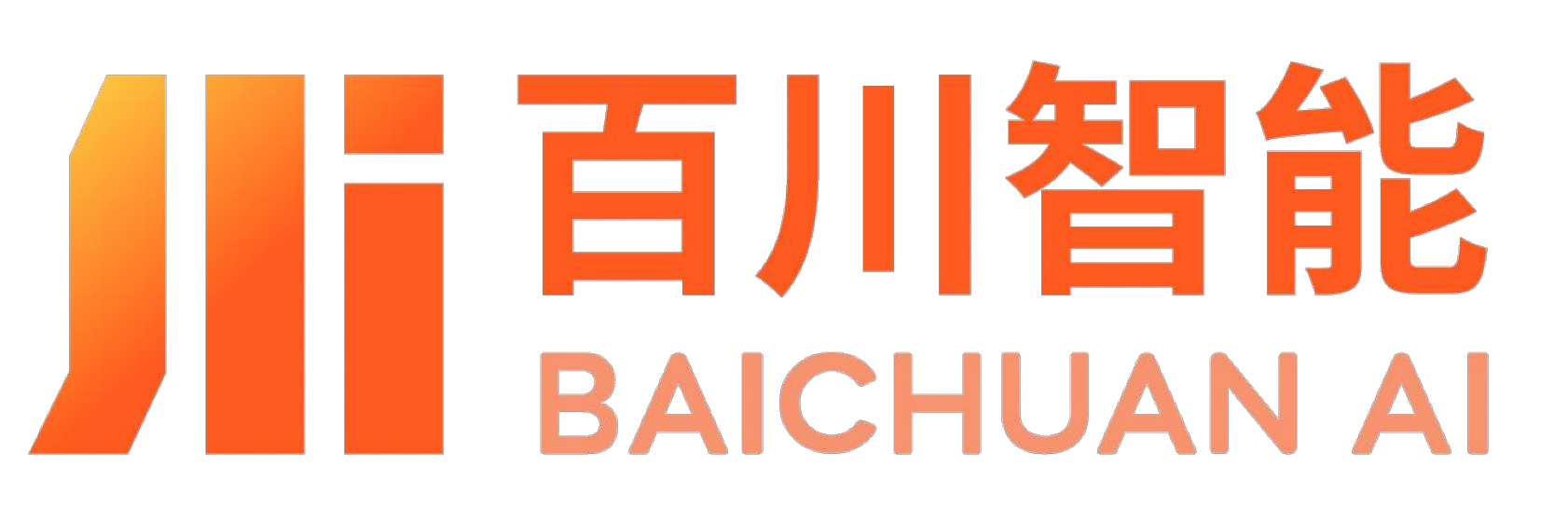}}
    \fancyhead[R]{\footnotesize\itshape Baichuan-M3 Technical Report}
    \fancyfoot[C]{\thepage}
    
}

\fancypagestyle{fancy}{
    \fancyhf{} 
    \fancyhead[L]{\includegraphics[height=\logoheight]{\logofile}}
    \fancyhead[R]{\footnotesize\itshape Baichuan-M3 Technical Report}
    \fancyfoot[C]{\thepage}
    
}

\title{Baichuan-M3: Modeling Clinical Inquiry for Reliable Medical Decision-Making}

\author{\href{https://www.baichuan-ai.com/}{\normalsize Baichuan-M3 Team}}

\usepackage{array}

\AtBeginBibliography{\sloppy\emergencystretch=1em\raggedright}

\begin{document}

\setlength{\emergencystretch}{1em}

\maketitle

\begin{abstract}
  We introduce Baichuan-M3, a medical-enhanced large language model engineered to shift the paradigm from passive question-answering to active, clinical-grade decision support. Addressing the limitations of existing systems in open-ended consultations, Baichuan-M3 utilizes a specialized training pipeline to model the systematic workflow of a physician. Key capabilities include: (i) proactive information acquisition to resolve ambiguity; (ii) long-horizon reasoning that unifies scattered evidence into coherent diagnoses; and (iii) adaptive hallucination suppression to ensure factual reliability. Empirical evaluations demonstrate that Baichuan-M3 achieves state-of-the-art results on HealthBench, the newly introduced HealthBench-Hallu and ScanBench, significantly outperforming GPT-5.2 in clinical inquiry, advisory and safety. The models are publicly available at \href{https://huggingface.co/collections/baichuan-inc/baichuan-m3}{https://huggingface.co/collections/baichuan-inc/baichuan-m3}.  
\end{abstract}

\section{Introduction}
Large language models (LLMs) are advancing rapidly~\cite{gpt4, o1, bc2,omni15}, driving broader adoption in healthcare~\cite{bcm1,baichuan-m2,MedGemma,Lingshu,Hulu-Med}. As a result, expectations are shifting from one-off question answering to end-to-end clinical decision support~\cite{MedDM,ArgMed}. This trend is reflected in leading systems such as OpenAI’s GPT-5.2~\cite{openai2025gpt52}, ChatGPT Health~\cite{chatgpt_health}, and Claude in Healthcare~\cite{anthropic_healthcare}. Yet key limitations persist: despite better scores on static, well-specified benchmarks, models often fail to stay evidence-grounded and uncertainty-aware in open-ended clinical interactions, where missing information and long-horizon decisions make hallucinations harder to control~\cite{hallubug}. Our goal is therefore to move beyond reliable QA toward decision-support partners that can operate safely in practice.

To bridge the gap between passive retrieval and active clinical support, recent studies have been evaluated along two partially disconnected paradigms: (i) factuality-focused, single-turn benchmarks and (ii) process-oriented, multi-turn consultation simulations. Benchmarks such as HealthBench~\cite{healthbench} and Med-HALT~\cite{ankit2023medhalt} measure factual consistency and common error modes, including hallucinations. They often show that performance drops on harder cases that require complex, multi-constraint clinical reasoning, and that models may produce ungrounded claims. In parallel, growing attention has been paid to Interactive History Taking (IHT) in OSCE-style~(Objective Structured Clinical Examination) settings, where models are assessed on their ability to elicit missing information and follow clinical workflows. Systems such as Google’s AMIE~\cite{tao2024amie, elahe2025amie} demonstrate strong communication quality in simulated encounters under these rubrics. 

However, a critical gap remains in unifying these two dimensions. Existing approaches often treat conversational interaction and clinical reasoning as orthogonal objectives, rather than components of a single coherent system. Knowledge-centric models exhibit “inquiry inertia,” lacking the agency to elicit missing evidence, while interaction-focused models can sacrifice diagnostic depth, prioritizing fluency over principled differential reasoning.

Integrating interaction and reasoning is challenging due to three technical bottlenecks. First, heterogeneous training environments across diverse clinical tasks hinder stable multi-task fusion. Second, long-horizon interactions amplify the credit-assignment problem in reinforcement learning (RL)~\cite{guo2025deepseek,bcalign,DBLP:conf/cacre/CaoZ19}: when supervision is dominated by terminal outcomes, models struggle to identify which conversational turns were causally responsible for diagnostic success. Third, efforts to increase reasoning depth often encounter reward saturation, where learning signals diminish near performance plateaus—sometimes accompanied by increased hallucination as models attempt to satisfy complex logical constraints~\cite{DBLP:journals/corr/abs-2505-23646,DBLP:journals/corr/abs-2510-22977}.

To address these challenges, we introduce Baichuan-M3, a next-generation medical LLM designed to unify clinical inquiry with reliable decision-making. Baichuan-M3 emphasizes three core competencies aligned with real clinical workflows: (i) proactive information acquisition, (ii) construction of coherent reasoning trajectories, and (iii) adaptive hallucination suppression.

Methodologically, we propose a three-stage training framework consisting of Task-Specific Reinforcement Learning (TaskRL), Offline Policy Distillation, and Multi-Teacher Online Policy Distillation (MOPD). This hierarchical design decouples the optimization of individual competencies before integrating them into a single policy. To improve long-horizon consultation performance, we introduce Segmented Pipeline Reinforcement Learning, which decomposes complex tasks into stages with separate reward signals. We further strengthen diagnostic reasoning via Dynamic Rubric Evolution and targeted RL objectives for hallucination suppression.

Baichuan-M3 achieves state-of-the-art performance on three authoritative benchmarks: (1) 44.4 on HealthBench-Hard, outperforming GPT-5.2; (2) top performance across all three dimensions of ScanBench (our OSCE-like benchmark)—Clinical Inquiry (74.9), Laboratory Testing (72.1), and Diagnosis (74.4)—exceeding both GPT-5.2-High and expert baselines; and (3) superior factual reliability in tool-free hallucination assessments.
In summary, our contributions are three-fold: 
\begin{itemize}
    \item \textbf{Bridging Inquiry and Reasoning:} We present Baichuan-M3, designed to transform LLMs from passive information retrievers into robust decision-support partners. By modeling the full clinical decision-making process, our system demonstrates the agency to actively elicit missing data while maintaining rigorous diagnostic logic, addressing the critical limitation of "hallucination via assumption" in open-ended scenarios.

    \item  \textbf{Clinical-Process-Aligned Optimization:} We introduce a training paradigm that mirrors professional medical training, utilizing Segmented Pipeline RL to align model behavior with distinct stages of clinical consultation (inquiry, lab testing, diagnosis). This approach, combined with dynamic rubric evolution, ensures that the model learns to prioritize evidence-based reasoning over mere conversational fluency or forced logical fitting.

    \item \textbf{Superior Empirical Results:} Extensive experiments demonstrate Baichuan-M3's leadership in both factual reliability and interactive procedural rigor. It sets new state-of-the-art records on HealthBench-Hard and our proposed ScanBench, showing significant gains in hallucination suppression and diagnostic accuracy compared to leading proprietary models like GPT-5.2 and expert baselines.
\end{itemize}

\section{Training Infrastructure}
In this section, we describe the training infrastructure of Baichuan-M3, including a stable patient simulation environment, a verification system that integrates rubric-based and fact-aware evaluation, and a progressive multi-stage training pipeline. Together, these components provide reliable interaction signals and scalable optimization support for long-horizon medical training.

\subsection{Patient Simulator}
In our previous work~\cite{baichuan-m2}, we introduced a patient simulator for doctor–patient interactions. During deployment, we found that the simulator became unstable when modeling proactive patients. Specifically, such behaviors disrupt the consultation flow, leading to simulations difficult to reproduce. To balance the generalization benefits of stochasticity with the stability required for long-horizon training, we adopt a passive-personality patient simulator and implement two complementary script-driven modes, referred to as Passive Interaction Mode and Interruption-Injected Mode.

\paragraph{Passive Interaction Mode (75\% sampling probability):}
This mode provides only the patient profile, inquiry rubrics, and behavioral constraints rubrics, without any predefined dialogue history. It simulates the opening phase of an initial consultation. Starting from an empty interaction state, this mode evaluates the physician agent’s ability to proactively elicit relevant information and form an initial diagnostic hypothesis under high uncertainty.

\paragraph{Interruption-Injected Mode (25\% sampling probability):}
This mode augments the basic patient information with a predefined dialogue snippet to simulate a mid-consultation state. The snippet ends with a patient-initiated question (often anxiety-driven about severity or treatment), modeling a common interruption during inquiry. However, exposing this snippet to the patient simulator may cause it to mimic the snippet’s speaking style and deviate from our passive-response protocol, introducing instability. We therefore use an asymmetric visibility mechanism: the snippet is visible only to the physician agent and hidden from the patient simulator. The physician answers the question and continues the consultation, while the simulator passively receives and responds to subsequent queries.

To reduce distribution mismatch between the simulated environment and real-world online consultations, we further introduce fine-grained probabilistic configurations in Interruption-Injected Mode. Specifically, end-of-turn questioning accounts for $50$\% of cases, where the patient question appears only at the end of the snippet, testing the model’s ability to handle abrupt interruptions. The remaining $50$\% corresponds to mid-turn questioning, which injects questions within an ongoing turn to simulate multi-turn. This mixed configuration ensures robust diagnostic performance under interaction noise, including frequent interruptions, challenges, and anxiety-driven follow-up questions.

\subsection{Verify System}
To guarantee the reliability and clinical safety of the generated responses, we construct a comprehensive Verify System that serves as the primary source of reward signals for Reinforcement Learning. Unlike general-domain chat models where fluency and helpfulness are often sufficient, medical agents face the dual challenge of strictly adhering to diagnostic protocols while ensuring absolute factual precision. A monolithic reward model often struggles to disentangle these orthogonal dimensions—potentially encouraging fluent hallucinations over rigid accuracy. To address this, our system decouples the evaluation process into two parallel streams: a \textbf{Rubric Verifier} that assesses the structural quality and adherence to clinical guidelines through fine-grained criteria, and a \textbf{Fact Verifier} that rigorously checks the biological and medical validity of atomic claims against external authoritative sources. This hybrid approach ensures the model is optimized for both professional compliance and factual groundedness.

\subsubsection{Rubric Verifier}
The M3 verification stack follows the rubric-based verifier paradigm~\cite{DBLP:journals/corr/abs-2508-12790,DBLP:journals/corr/abs-2507-17746} introduced in M2~\cite{baichuan-m2}. Instead of treating medical response quality as a single monolithic preference signal, we decompose each interaction into a set of independently decidable rubric clauses. An LLM-based judge~\cite{DBLP:journals/corr/abs-2411-15594} evaluates each clause item by item, and the resulting decisions are aggregated into a scalar reward for policy optimization.


Given a sample $x$ and a set of rubrics $\mathcal{R}=\{r_i\}_{i=1}^N$, each rubric $r_i$ is assigned a signed weight $w_i\in[-10,10]$, where $w_i>0$ denotes rewards and $w_i<0$ denotes penalties. An LLM judge outputs a binary decision $a_i\in\{0,1\}$ indicating whether $r_i$ is satisfied. The task reward is obtained by min-max normalization:
\begin{equation}
R_{\text{task}}
=\frac{\sum_{i=1}^N w_i a_i-\sum_{i:w_i<0} w_i}{\sum_{i=1}^N |w_i|}
\end{equation}

This normalization decouples the reward scale from both the number of rubrics and the magnitude of their weights into $[0,1]$. As a result, reward distributions remain comparable across different samples and rubric configurations, while integer weights can be used to directly encode relative clause importance in an interpretable manner.

To increase the efficiency in the RL pipeline, reward evaluation is scheduled asynchronously with the rollout process: once the policy emits a response, we immediately launch the corresponding rubric-judging jobs, overlapping judge compute with other unconflicted RL actions (e.g., computing $\pi_\text{ref}/\pi_\text{old}$ log-probabilities) to accelerate the training. In addition, we adopt a prefix-affinity prompt design for the judge: system constraints, output schema, and dialog context share the same template prefix, while the suffixes are substituted with the rubric clause. This processing performs micro-batching to maximize KV-cache~\cite{DBLP:journals/tmlr/0002LTTXCHD0025} reuse, substantially reducing the time cost. 

\subsubsection{Fact Verifier}
\label{sec:fact-verifier}
Baichuan-M2~\cite{baichuan-m2} mainly relies on rubric-based rewards to shape medical reasoning. As performance saturates on common criteria, the model tends to chase marginal gains by adding rarer clinical details, which increases hallucination risk; we therefore optimize for both completeness and reliability.
To make the objective of a low hallucination rate quantitative and optimizable, we build upon existing long-form factuality evaluation methodologies~\cite{DBLP:conf/emnlp/MinKLLYKIZH23,wei2024long} to construct a \textbf{Fact-Aware Verification Pipeline}. The pipeline adopts a two-stage architecture: Firstly, we decompose long-form responses into fine-grained, independently verifiable atomic claims. Secondly, we employ a search-augmented verification agent to validate each claim against authoritative sources. This module runs in parallel with the Rubric Verifier as a service, minimizing latency.

\paragraph{Atomic Claim Extraction Model.}

The foundation of the pipeline is transforming unstructured model outputs into discrete, verifiable units, referred to as atomic claims. These claims must be self-contained and verifiable even when detached from their original context. To meet this requirement, we apply the following rules:

\begin{enumerate}
    \item \textbf{Atomicity and Coreference Resolution.} We decompose complex compound sentences into single-fact units and resolve pronouns to guarantee semantic independence.
    \item \textbf{Noise and Distractor Filtering.} We discard units that lack sufficient context to form a factual statement. Crucially, in multiple-choice scenarios, we explicitly exclude the recitation of incorrect options (distractors) to prevent them from being misclassified as hallucinations.
    \item \textbf{Deduplication with Order Preservation.} Semantically redundant assertions are eliminated while maintaining the original logical sequence of the reasoning chain.
\end{enumerate}

In the evaluation phase, we initially use GPT-5~\cite{singh2025openai} as a high-quality extractor; however, its inference latency is prohibitive for online RL. We therefore distill an efficient 8B extraction model from GPT-5, trained on datasets spanning multiple medical subtasks, and use it as the final extractor, with detailed fidelity evaluation provided in Appendix \ref{sec:extraction_model_evaluation}.

\paragraph{Search-Augmented Verifier.}

After extraction, atomic claims are fed into a search-augmented validation agent. This agent performs iterative searches over authoritative medical sources such as clinical guidelines and autonomously decides whether the collected evidence is sufficient to support a verdict. Each claim is finally assigned one of three labels: Supported, Refuted or, Uncertain. Compared with methods that rely only on parametric model knowledge or static knowledge bases, this dynamic search-based mechanism can stay aligned with the latest clinical evidence and better handle the continuous evolution of medical knowledge.

\paragraph{Two-Level Caching System for Acceleration.}

Introducing fine-grained fact verification into the online RL loop raises significant computational challenges. A single RL iteration may generate thousands of atomic claims, and performing real-time external search for each claim is unacceptable in both cost and latency.

To accelerate the training, we design a two-level claim caching system based on the key observation that, for the same clinical query, multiple sampled responses from the model differ in wording but share a high proportion of underlying medical claims.

\begin{itemize}
    \item \textbf{Level-1 cache (exact match).} We use Redis to cache verification results for identical claim strings, enabling millisecond-level lookups and result reuse.
    \item \textbf{Level-2 cache (semantic match).} We store embeddings of historical claims in a vector database and apply ANN retrieval to find semantically equivalent claims and reuse their verification results.
\end{itemize}

As the cache pool grows, the overall hit rate increases from below 40\% in the early stage to around 80\%. This reduces external search requests by approximately 85\%, making the impact of fact verification on training is negligible. Semantic caching inevitably introduces some systematic bias. For example, claims with subtle dosage differences may be incorrectly treated as equivalent. We address this issue using the signal denoising mechanism described in Section~\ref{sec:fact-aware}.

\subsection{Multi-Task Training Pipeline}

\begin{figure}
    \centering
\includegraphics[width=1.0\linewidth]{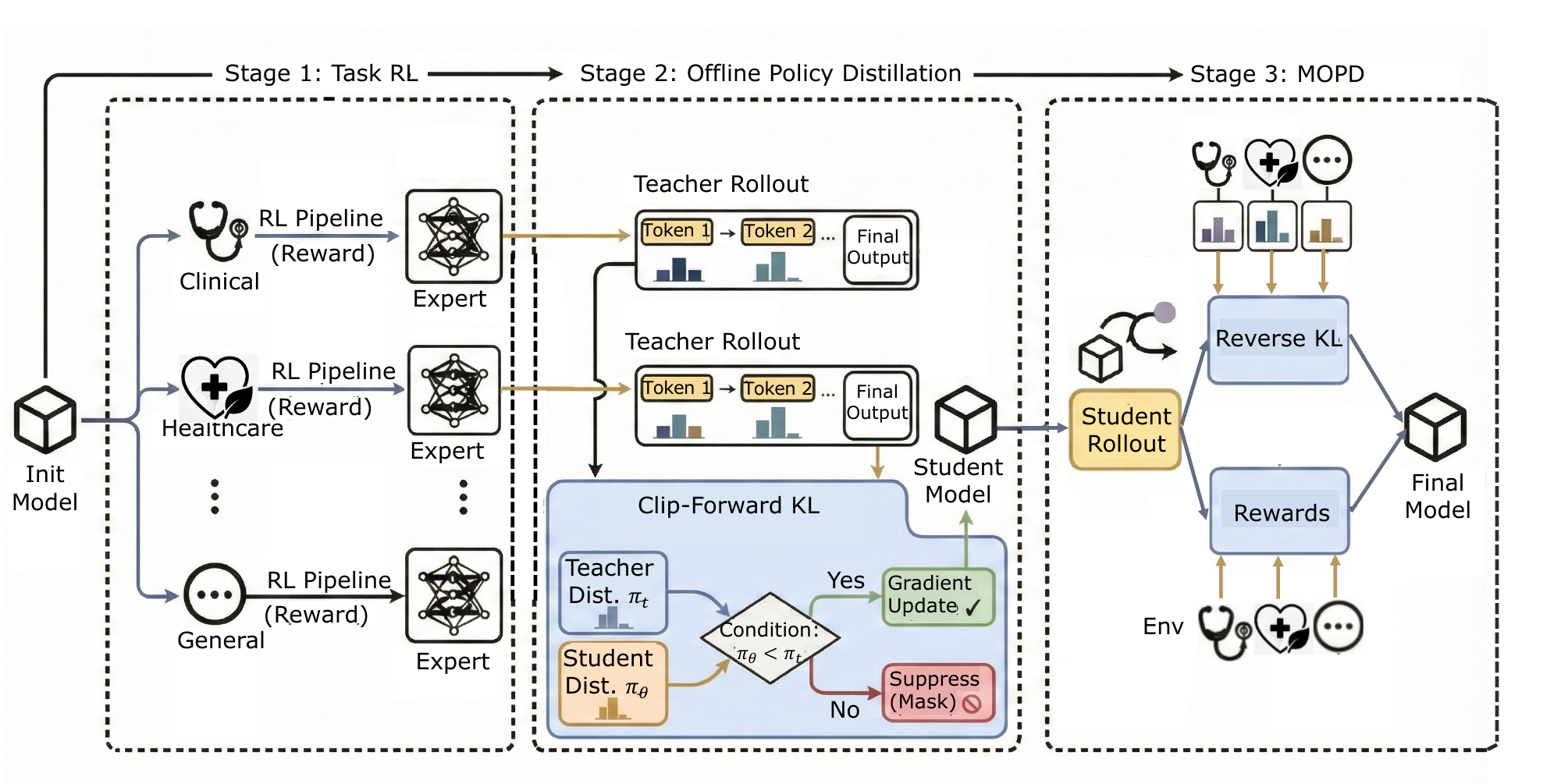}
    \caption{An illustration for the three stage pipeline.}
    \label{fig:overall}
\end{figure}

To mitigate the trade-off issue in multi-task learning~\cite{DBLP:journals/corr/abs-2007-10527,DBLP:conf/icassp/MoW22} and reduce development complexity, the training pipeline of Baichuan-M3 is decomposed into three progressive stages: Capability Learning, Distribution Fusion, and Policy Unification. By isolating the acquisition of expert capabilities from the student models, this design achieves a stable fusion process and significantly improves the overall performance. An illustration for the overall pipeline is summarized in Fig.~\ref{fig:overall}.

\paragraph{Stage 1: Task RL.}

In the initial stage, our objective is to construct a diverse and high-quality set of expert teachers. Starting from a shared initialization, we deploy independent RL pipelines tailored to specific capability domains. Specifically, to address the diverse requirements from medical applications, we explicitly train specialized experts for Clinical Inquiry and Healthcare Consultation, ensuring the model captures the nuances of real-world diagnostic dialogue and patient-centric health advice. In addition, we also train a generalist expert focused on fundamental capabilities, including Instruction Following and General Reasoning.

Inspired by \cite{DBLP:journals/corr/abs-2510-13361}, the core philosophy here is differentiation rather than unification. By allowing different models to explore fully under their respective task-specific rewards, we obtain a set of domain-specialized teachers with strong, distinct inductive biases. This divide-and-conquer strategy following \cite{DBLP:journals/corr/abs-2510-13361} effectively isolates gradient interference across tasks, avoiding the optimization conflicts typical in the early stages of multi-task mixture training, thereby providing high-confidence behavioral guidance for subsequent fusion.

\paragraph{Stage 2: Offline Policy Distillation.}
The second stage focuses on ``compressing'' the capabilities of multiple teachers into a single student model via offline distillation. We freeze all teacher models and perform rollouts within their respective domains to construct an offline trajectory dataset, $\mathcal{D}$. The student model then learns from this data in an off-policy manner.

To adapt to the reality of single-sample offline data and ensure numerical stability during fusion, we introduce a standard KL divergence in favor of a restricted distillation objective based on Clip-Forward-KL. For each sample $(s, a)\in\mathcal{D}$ and its corresponding teacher policy $\pi_t$, we define the loss function as follows:

\begin{equation}
\mathcal{L}_{\text{clip-FKL}}(\theta) = \mathbb{E}_{(s, a) \sim \mathcal{D}} \left[ \mathbb{I} \left( \log \pi_\theta(a|s) < \log \pi_t(a|s) \right) \cdot \left( - \log \pi_\theta(a|s) \right) \right]
\end{equation}

where $\mathbb{I}(\cdot)$ denotes the indicator function and $\pi_\theta(\cdot)$ and $\pi_t(\cdot)$ denotes the token-wise probability of student and teacher model, respectively. This design applies a one-sided update that only enforces non-inferiority on the teacher’s empirical support, rather than overfitting to the full conditional distribution. This avoids probability over-amplification in the single-sample regime and implicitly preserves entropy outside the data support. It can mitigate the mode collapse and leave exploration space for the subsequent mode-seeking optimization in Stage 3.

Leveraging the mode-covering property of Forward KL, this stage enables the student model to broadly cover high-probability regions of different expert distributions. This facilitates the stable inheritance of behavioral patterns and eliminates the instability brought by cold-start associated with direct multi-objective RL.

\paragraph{Stage 3: Multi-Teacher On-Policy Distillation (MOPD)~\cite{mimo2025flash}.}
In the third stage, the student model re-enters the online interaction environment, performing rollouts across mixed domain distributions. At this point, the model is constrained simultaneously by ground-truth task rewards and multi-teacher priors. Unlike the distribution imitation in the second stage, this stage employs reverse KL regularization~\cite{lu2025onpolicydistillation}.

Leveraging the mode-seeking nature of Reverse KL, the student is driven to select the optimal mode when comprised with conflicting advice from multiple teachers, rather than passively averaging them. Guided by real reward signals, the student transitions from an ``imitator'' to a ``decision-maker,'' achieving deep unification and de-noising capability at the policy level.

\paragraph{The Evolution of Teacher Capability.}
 Our proposed framework is not a static pipeline and supports cyclic iterative refinement. The unified model obtained after MOPD can serve as a new initialization for the Stage 1 to achieve domain-specific enhancement, followed by another round of distillation. This flexibility allows for continuous improvement of the model's capabilities with low marginal cost.

\section{Task-specific Training Methods}
Clinical practice encompasses a diverse spectrum of cognitive activities, ranging from the active elicitation of symptom history to the rigorous provision of evidence-based advice. A monolithic training approach often fails to balance the distinct requirements of these scenarios—specifically, the deductive logic required for diagnosis versus the strict factual adherence required for advisory services. To address this, we adopt task-specific optimization strategies tailored to the unique modalities of medical interaction. In this section, we present our specialized methodologies for two core capabilities: Deep Clinical Consultation, where we employ a segmented pipeline and the Step-Penalized Advantage with Relative baseline algorithm to master the multi-turn diagnostic trajectory; and Credible Healthcare Advisory, which utilizes dynamic rubric evolution and Fact-Aware Reinforcement Learning to ensure the safety and verifiability of medical information.

\begin{figure}[b]
    \centering
    \includegraphics[width=0.98\linewidth]{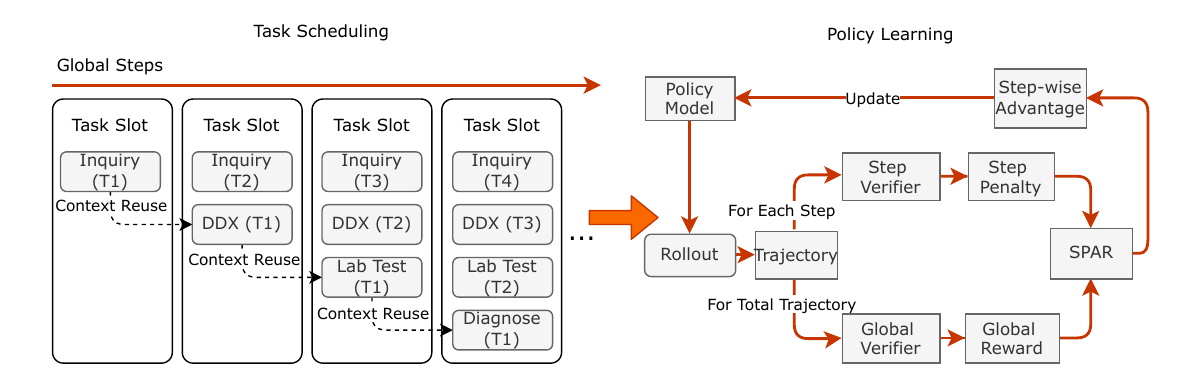}
    \caption{Segmented Pipeline RL (left) and Policy Learning Algorithm (right).}
    \label{fig:spar}
\end{figure}

\subsection{Deep Clinical Consultation}
As medical AI evolves, the demand for "instant diagnosis based on symptom input" has surged~\cite{liu2025medchainbridginggapllm}. However, clinical medicine transcends simple knowledge retrieval; it is a discipline grounded in rigorous evidence and deductive logic. Because identical symptoms may originate from diverse etiologies and patient-specific risk thresholds vary, reliable clinical decisions must rely on granular data, systematic risk assessment, and traceable reasoning~\cite{liao2020task, dou2024plpf}.

We introduce the Deep Clinical Consultation framework, which re-imagines the consultation as a clinical-grade, structured, and auditable process of information production. Our objective is to collect pivotal clinical data within brief interactions while maintaining stringent safety standards. To achieve this, we propose a training framework (see Fig.~\ref{fig:spar}) comprising a Segmented Pipeline RL architecture and the Step-Penalized Advantage with Relative baseline (SPAR) algorithm.

\subsubsection{Segmented Pipeline RL}
We formulate the consultation as a $K$-stage generation process with $K=4$. Let $[K]=\{1,\dots,K\}$ index the stages, and let $\mathcal{S}=\{\text{Inq},\text{DDX},\text{Lab},\text{Diag}\}$ denote the set of stage types.
For a patient case $i$ at stage $k$, let $x_{k}^{(i)}$ denote the current input context, which encapsulates the entire trajectory history plus the specific instruction for the current stage (e.g., "Based on the above, suggest lab tests"). The policy $\pi_\theta$ generates a response segment $y_{k}^{(i)}$:
\begin{equation}
    y_{k}^{(i)} \sim \pi_\theta(\cdot | x_{k}^{(i)})
\end{equation}
As shown in Fig.~\ref{fig:spar}, we employ an Asynchronous Multi-Task Pipeline in which, at global step $t$, multiple task slots run in parallel on different pipeline stages (and potentially different patient cases); the union of these stage-specific tasks forms the training batch $\mathcal{B}_t$.

\paragraph{Multi-Task Training Objectives.}
The optimization targets the quality of the generated segment $y_k$ relative to the stage-specific goals. For a batch of active contexts $\mathcal{B}_t = \{x_{k_i}^{(i)}\}_{i=1}^N$, the joint objective function is:
\begin{equation}
    \mathcal{J}(\theta) = \frac{1}{|\mathcal{B}_t|} \sum_{x_{k}^{(i)} \in \mathcal{B}_t} \mathcal{L}_{\text{RL}} \left( y_{k}^{(i)}, \mathcal{G}_{k}^{(i)} \right)
\end{equation}
where $\mathcal{G}_k$ is the reward function tailored to stage $k$. This formulation unifies diverse reasoning capabilities (gathering vs. deducing) into a single generative model.

\paragraph{Context Reuse via Gated Transition.}
The pipeline relies on the auto-regressive accumulation of context. The input for the next stage, $x_{k+1}^{(i)}$, is constructed by appending the generated response $y_{k}^{(i)}$ and the next stage's instruction $p_{k+1}$ to the current context:
\begin{equation}
    x_{k+1}^{(i)} = [ x_{k}^{(i)}, y_{k}^{(i)}, p_{k+1} ]
\end{equation}
However, open-ended generation carries the risk of error propagation. To enforce the "garbage-in, garbage-out" principle, we implement a Quality-Gated Transition. The training pool for the subsequent stage, $\mathcal{D}_{k+1}$, is populated via a filtration process:
\begin{equation}
    \mathcal{D}_{k+1} \leftarrow 
    \begin{cases} 
        \mathcal{D}_{k+1} \cup \{ [ x_{k}^{(i)}, y_{k}^{(i)}, p_{k+1} ] \}, & \text{if } \mathcal{V}_{k}^{(i)} \ge \tau \\
        \mathcal{D}_{k+1}, & \text{otherwise (Discard)}
    \end{cases}
\end{equation}
where $\mathcal{V}_{k}^{(i)}$ is the score produced by the stage-$k$ quality verifier and $\tau$ is the acceptance threshold. This ensures that only trajectories with clinically valid logical chains are extended, effectively pruning error paths from the training curriculum.

\subsubsection{Policy Learning Algorithm: SPAR}
Conventional RL approaches, such as GRPO~\cite{guo2025deepseek, DBLP:journals/corr/abs-2509-02333} with global rewards, exhibit significant limitations in long-horizon medical interviewing. These include reward hacking (inflating recall via redundant questions), logic fragmentation (disjointed clinical transitions), and ineffective credit assignment. In long-context dialogues, trajectory-level rewards fail to isolate local errors~\cite{zhang2025lets}, often inducing training instability by penalizing valid reasoning chains alongside specific flaws.

To address these challenges, we propose SPAR (Step-Penalized Advantage with Relative baseline), which introduces fine-grained step-wise penalties and a decoupled advantage estimation mechanism to induce an adaptive curriculum-learning effect.

\paragraph{Reward Formulation.}
SPAR employs a hierarchical reward structure. For a generated response $y$ consisting of $L$ logical interaction steps $[z_1, \dots, z_L]$, we define each step $z_j$ as one complete dialogue exchange (i.e., a user turn followed by the assistant turn). We first compute a Global Reward $R_{\text{global}}$ by evaluating the complete trajectory against a set of pre-defined Clinical Rubrics (e.g., diagnostic accuracy, evidence completeness).

Simultaneously, a step verifier performs real-time validation for each interaction step $z_j$. Let $\mathbb{V}_j$ denote the set of violation types triggered by step $z_j$ (e.g., redundancy, safety risk). We define the step-wise validity factor $\gamma_j \in (0, 1]$ as:
\begin{equation}
    \gamma_j =
    \begin{cases}
        1, & \text{if } \mathbb{V}_j = \emptyset \\
        \min\limits_{v \in \mathbb{V}_j} (\lambda_v), & \text{otherwise}
    \end{cases}
\end{equation}
where $\lambda_v \in (0, 1)$ is the penalty coefficient for violation type $v$. This formulation enforces a minimum validity principle: if a step commits multiple errors, only the most severe penalty (smallest $\lambda$) is applied. The effective return for the step is then modulated as $R_j = \gamma_j \cdot R_{\text{global}}$.

\paragraph{Step-wise Advantage Estimation.}
The core innovation of SPAR lies in its advantage computation, which decouples local penalties from the group baseline. Unlike traditional approaches that normalize rewards using the penalized distribution, SPAR computes the advantage $\hat{A}_{j}$ for step $z_j$ by comparing the \textit{step-penalized return} against an \textit{unpenalized group average}:
\begin{equation}
    \hat{A}_{j} = \frac{\gamma_{j} R_{\text{global}} - \mu_{\text{raw}}}{\sigma_{\text{raw}} + \epsilon}
\end{equation}
where $\mu_{\text{raw}}$ and $\sigma_{\text{raw}}$ represent the mean and standard deviation of the raw global rewards ($R_{\text{global}}$) within the sampled group, excluding any step penalties. Here, a sampled group refers to multiple rollouts generated under the same prompt (i.e., the same consultation context), which enables relative comparison among candidates.

We then apply a GSPO-style policy update~\cite{DBLP:journals/corr/abs-2507-18071}, using the step-wise advantages $\hat{A}_j$ to weight the likelihood-ratio objective, so that penalties are attributed to the specific interaction steps responsible for violations. Here, a sampled group refers to multiple rollouts generated under the same prompt (i.e., the same consultation context), which enables relative comparison among candidates.


\paragraph{Implicit Curriculum Mechanism.}
This advantage design facilitates an implicit curriculum~\cite{bengio2009curriculum} by naturally scheduling optimization priorities based on error severity:
\begin{itemize}
    \item \textbf{Phase 1: Correction of Critical Errors.} For severe violations (e.g., repetition), we assign a rigorous penalty (e.g., $\lambda \approx 0.1$). This forces the effective return $\gamma_j R_{\text{global}}$ significantly below the group baseline $\mu_{\text{raw}}$, resulting in a large negative advantage ($\hat{A}_j \ll 0$). This dominant gradient signal compels the model to prioritize rectifying fundamental usability flaws in the early training phase.
    \item \textbf{Phase 2: Refinement of Nuance.} For subtle imperfections (e.g., rigid phrasing), a milder penalty (e.g., $\lambda \approx 0.9$) is applied. Initially, this small deviation is masked by the high variance ($\sigma_{\text{raw}}$) of global rewards. However, as the policy stabilizes and $\sigma_{\text{raw}}$ decreases, these fine-grained signals begin to dictate the gradient direction, guiding the model toward stylistic perfection.
\end{itemize}
By isolating the impact of specific step-level behaviors, SPAR enables precise credit assignment, distinguishing local flaws from overall diagnostic success.

\subsection{Credible Healthcare Advisory}
This section details our training methodology for credible healthcare advisory, focusing on the synthesis of clinical credibility and interactive utility. To overcome the limitations of static feedback, we first introduce a dynamic rubric evolution framework designed to mitigate reward hacking and ensure that the model pursues genuine reasoning rather than superficial patterns. Furthermore, we present a Fact-Aware Reinforcement Learning strategy that enhances the model’s factual reasoning capability. By moving beyond naive penalty mechanisms, this approach effectively suppresses unfaithful hallucinations while circumventing induced conservatism, thereby preserving the model's ability to provide detailed and informative medical counsel.

\subsubsection{Dynamic Rubric Evolution}
In our previous work~\cite{baichuan-m2}, we proposed a rubric-based RL approach to enhance clinical reasoning capabilities. However, we observed that this method is highly susceptible to ``reward hacking.'' The model tends to pursue superficial ``high-score performance'' rather than genuine reasoning, manifesting as passive verbosity, defensive template usage, or hallucinated details. The root cause lies in the fact that prior rubric formulation relied solely on the \textit{input question}. Once the question is fixed, the rubric becomes static, creating a predictable structure that the model can easily exploit.

To address this issue, we introduce a significant upgrade to the fundamental quality of the feedback signal in the M3 model, compared to its predecessor, M2. Inspired by \cite{DBLP:journals/corr/abs-2511-19399,DBLP:journals/corr/abs-2509-21500}, we propose a Human-AI Collaborative Dynamic Evolution mechanism, which crucially incorporates the \textit{model's response} into the rubric synthesis process. Specifically, we categorize constraints into two distinct classes:

\begin{itemize}
    \item \textbf{Core Rubric Set:} Synthesized solely based on the question, this set guides the general direction of model optimization and ensures fundamental safety.
    \item \textbf{Dynamic Rubric Set:} Synthesized dynamically based on both the question \textit{and} the model's historical responses. This set aims to constrain non-compliant behaviors and specific vulnerabilities discovered during the training process.
\end{itemize}
The construction and maintenance of this dynamic set rely on two key modules: Quality Control and Admission/Exit Rules.

\paragraph{Quality Control}
To ensure the efficacy and clinical relevance of the dynamically generated rubrics, we implement a ``Mine-Verify-Inject'' closed-loop workflow.
First, we identify High-Confidence Samples—responses that achieve high scores under the current system but harbor latent defects.
Next, the Rubric Mining Agent analyzes these samples to identify adversarial patterns and draft candidate constraints.
Finally, Human Experts intervene to validate these candidates. Instead of authoring rules from scratch, experts review the agent's output, assessing its boundary determinism and compliance with meta-principles (e.g., Safety $>$ Empiricism). This ensures that the rubric incentivizes reasoning rather than shifting the locus of reward hacking.

\paragraph{Admission and Exit Rules}
To prevent rule explosion and ensure the reward signal remains potent, the dynamic rubric set follows a ``Problem-Driven'' lifecycle:

\begin{itemize}
    \item \textbf{Admission:} A candidate rubric is not admitted merely for its validity; it is activated only when it targets a statistically significant failure mode (i.e., a high violation rate in model responses). This ensures feedback remains focused on active behavioral deficits.
    \item \textbf{Exit:} Once a constraint is consistently satisfied over multiple training epochs (violation rate $\rightarrow 0$), it is automatically retired from the dynamic set. This ``pruning'' mechanism prevents reward signal dilution, ensuring the optimization focus remains strictly on emerging and unresolved issues without being washed out by redundant positive rewards.
\end{itemize}

\subsubsection{Fact-Aware Reinforcement Learning}
\label{sec:fact-aware}

\begin{figure}[t]
    \centering
    \includegraphics[width=0.98\linewidth]{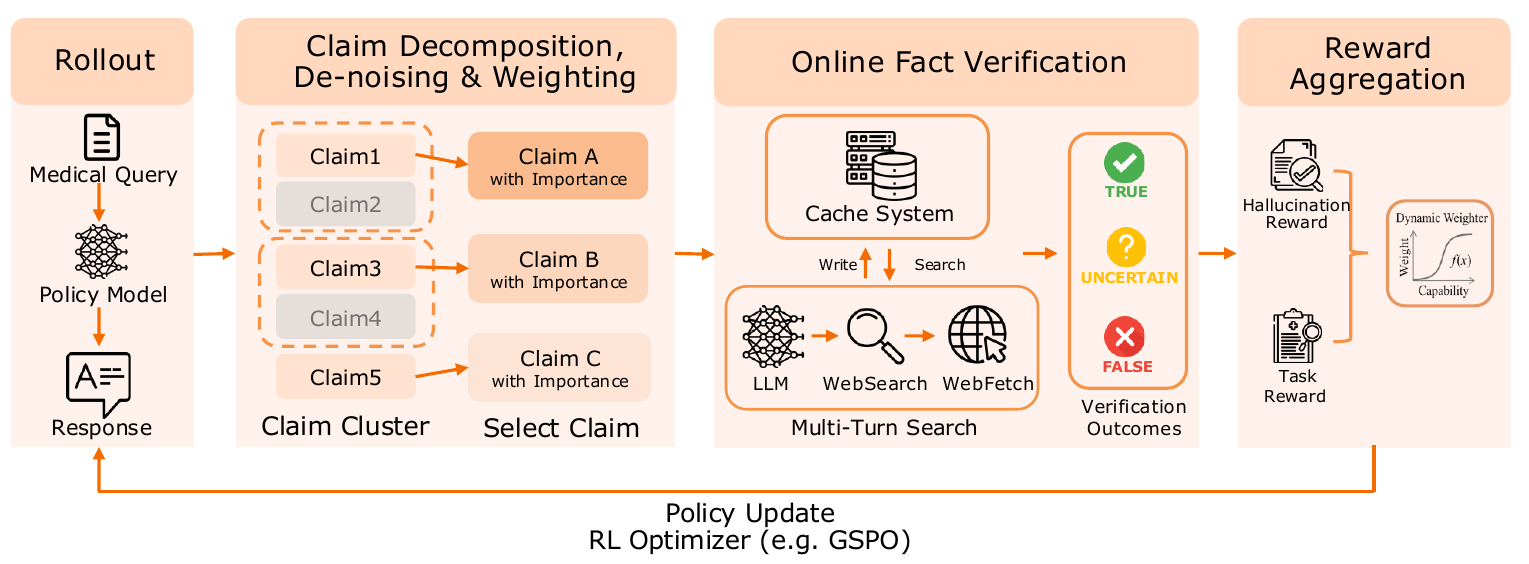}
    \caption{Fact-Aware Reinforcement Learning Algorithm.}
    \label{fig:fact_aware_rl_fig}
\end{figure}

In the paradigm of Reinforcement Learning with Verification Rewards (RLVR), naive hallucination suppression strategies typically convert verification signals directly into scalar penalties~\cite{DBLP:conf/emnlp/MinKLLYKIZH23, chen2025learning}. The standard objective form is defined as
\begin{equation}
R = R_{task} + \alpha \cdot R_{hallu}
\label{eq:naive-objective}
\end{equation}
Here, $R_{hallu}$ utilizes a count-based hallucination rate metric ($-N_{\text{hallu}}/N_{\text{total}}$). While this objective aims to preserve core medical reasoning capability ($R_{task}$) while reducing hallucinations via a penalty term, it is vulnerable to two major forms of reward hacking in long-form generation:

\begin{itemize}
    \item \textbf{Redundancy-Induced Dilution:} The model inflates the denominator $N_{\text{total}}$ by producing many factually correct but low-value statements, thereby reducing the measured hallucination rate without correcting the core errors.

    \item \textbf{Penalty-Induced Conservatism:} Strict penalties can encourage overly conservative strategies (e.g., shortening outputs to avoid penalties), which undermines exploration and the acquisition of complex reasoning behaviors.
\end{itemize}
To address these issues, we propose a joint optimization framework comprising Structured Signal Denoising and Dynamic Multi-Objective Aggregation, as shown in Fig.~\ref{fig:fact_aware_rl_fig}.

\paragraph{Structured Signal Denoising}

To establish a reward signal robust to redundancy, we reformulate the verification of atomic claims as a weighted evaluation based on semantic density. This ensures that hallucinations in core diagnostic statements incur significantly higher penalties than those in marginal content.

We first map the response sequences $\{s_j\}$ and extracted claims $\{c_i\}$ into a vector space via a semantic encoder $\mathcal{E}(\cdot)$. To prevent the manipulation of metrics through synonymous paraphrasing, we apply semantic clustering with a cosine similarity threshold. By selecting a representative claim $c_k^*$ for each cluster $\mathcal{C}_k$, we transition the evaluation metric from lexical frequency to semantic unit density. We then quantify the information contribution of each claim using a Saliency Weight, $w(c_k^*)$, defined as its maximum semantic correlation across the response sentences:
\begin{equation}
w(c_k^*) = \max_{1 \le j \le M} \cos(\mathcal{E}_{c_k^*}, \mathcal{E}_{s_j})
\end{equation}
The factuality reward $R_{fact}$ is computed as a weighted penalty term:
\begin{equation}
R_{fact} = - \frac{\sum_{k=1}^{K} w(c_k^*) \cdot \mathbb{I}(c_k^*)}{\sum_{k=1}^{K} w(c_k^*) + \epsilon}
\end{equation}
where $\mathbb{I}(c_k^*)$ is the verification penalty indicator defined as:
\begin{equation}
\mathbb{I}(c_k^*) = 
\begin{cases} 
1 & \text{if } c_k^* \in \{\text{Refuted}, \text{Uncertain}\} \\
0 & \text{otherwise}
\end{cases}
\end{equation}
Mechanically, the weighted denominator neutralizes the inflation of claim counts (anti-dilution), while the saliency-dependent numerator ensures that penalties are concentrated on core errors rather than marginal text, thereby preserving reasoning utility.

\paragraph{Dynamic Multi-Objective Aggregation}

To balance medical reasoning reinforcement with hallucination suppression, we implement a dynamic aggregation mechanism. We introduce a soft-gating coefficient, $\lambda(R_{task})$, which modulates the penalty intensity based on the task reward achieved by the on-policy generated response.

Given the $R_{task}$ associated with a specific response, we compute the dynamic coefficient $\lambda(R_{task})$ strictly as a function of the task reward using a shaped Sigmoid function:
\begin{equation}
\lambda(R_{task}) = \sigma \left( \kappa \cdot \frac{R_{task} - \mu}{\Delta} \right)
\end{equation}
where $\sigma(\cdot)$ is the standard sigmoid function, centered at $\mu = (\tau_{min} + \tau_{max})/2$ and scaled by $\Delta = \tau_{max} - \tau_{min}$ (with steepness $\kappa=10$).

The thresholds are calibrated via posterior analysis of the task reward distribution. We specifically set $\tau_{min} = 0.75$ and $\tau_{max} = 0.95$ to delimit the critical interval reflecting effective medical reasoning. This setting aligns the penalty strength with the model's demonstrated capability:

\begin{itemize}
    \item \textbf{Protection Zone ($R_{task} < \tau_{min}$):} $\lambda(R_{task}) \to 0$. Penalties are suppressed to prioritize capability optimization, shielding the acquisition of fundamental reasoning skills from interference.
    \item \textbf{Transition Zone ($\tau_{min} \le R_{task} \le \tau_{max}$):} $\lambda(R_{task})$ increases non-linearly. In this phase, the system progressively introduces factual constraints.
    \item \textbf{Constraint Zone ($R_{task} > \tau_{max}$):} $\lambda(R_{task}) \to 1$. Full penalties are enforced to maximize the suppression of hallucinations once the model demonstrates sufficient reasoning competence.
\end{itemize}

The final total reward $R$ aggregates the task utility with the gated factuality penalty:
\begin{equation}
R = R_{task} + \lambda(R_{task}) \cdot R_{fact}
\label{eq:full-fact-aware-rl}
\end{equation}
This formulation implements an implicit curriculum that secures reasoning competence before imposing rigorous safety constraints. We empirically validate this behavior through comparative ablation studies in Appendix \ref{sec:ablation_reward}.



\section{Evaluation}
To comprehensively evaluate the clinical utility and safety of Baichuan-M3, we conduct rigorous experiments across two complementary dimensions: dynamic clinical workflow simulation (\textbf{ScanBench}) and broad-spectrum medical reasoning (\textbf{HealthBench}~\cite{healthbench}). We benchmark Baichuan-M3 against a diverse set of competitive baselines to ensure a robust assessment. These include state-of-the-art general-purpose LLMs known for superior reasoning capabilities (e.g., \textbf{GPT-5.2-High}\cite{openai2025gpt52}, \textbf{Deepseek-V3.2-Thinking}\cite{deepseek2025v32thinking}, \textbf{Qwen3-235B-thinking-2507}\cite{qwen3-235b-thinking-2507}), representative medical-specific models (e.g., \textbf{AntAngelMed}\cite{antangelmed2026}), and our previous generation model (\textbf{Baichuan-M2}~\cite{baichuan-m2}). Crucially, to benchmark the model against real-world professional standards, we explicitly introduce a Human Baseline composed of attending physicians from Grade-A tertiary hospitals, each with a minimum of 5 years of clinical experience. This comparative analysis aims to verify the model's advancements in complex decision-making, specialized knowledge application, and hallucination suppression relative to generalist giants, domain specialists, and human experts.

\subsection{ScanBench}
Unlike traditional static question-answering benchmarks, ScanBench simulates the authentic clinical workflow of "Inquiry  $\rightarrow$ Lab Testing $\rightarrow$ Diagnosis." The dataset, slated for open-source release, transforms diverse clinical cases into an observable and quantifiable decision path divided into three progressive stages.

\subsubsection{Dataset Composition and Statistics}
ScanBench is constructed to simulate the authentic clinical environment with high fidelity across three dimensions: case diversity, inquiry granularity, and examination complexity.

\paragraph{Clinical Case Diversity}
As shown in Table~\ref{tab:scanbench_distribution}, ScanBench contains 303 cases from 12 departments. It covers both common conditions (e.g., General Practice) and long-tail specialties (e.g., Rheumatology, Hematology).

\begin{table}[H]
\centering
\caption{Distribution of Clinical Samples by Department.}
\label{tab:scanbench_distribution}
\begin{tabular}{lclc} 
\toprule
\textbf{Department} & \textbf{Count} & \textbf{Department} & \textbf{Count} \\ \midrule
General Practice    & 111 & Nephrology           & 15 \\
Surgery             & 50  & Cardiology           & 14 \\
Gynecology          & 26  & Respiratory Medicine & 14 \\
Neurology           & 25  & Endocrinology        & 7  \\
Gastroenterology    & 18  & Rheumatology         & 6  \\
Hematology          & 15  & Geriatrics           & 2  \\ \bottomrule
\end{tabular}
\end{table}

\paragraph{Inquiry Granularity and Rigor}
To quantify the inquiry process, we annotated 8,857 checklist items as ground truth.
\begin{itemize}
    \item \textbf{Information Density:} Each case contains 29.23 items on average (range: 20--35), requiring sustained multi-turn context tracking.
    \item \textbf{Logical Distribution:} The checklist mirrors clinical diagnostic logic: History of Present Illness dominates (55.8\%), followed by Past Medical History (19.6\%) and Personal/Social History (14.6\%), with Obstetric/Gynecological (5.4\%) and Family History (4.7\%) included when relevant.
    \item \textbf{Criticality Weighting:} We distinguish between essential safety points and general information: 51.3\% are labeled Level 2 (Critical) for diagnosis/risk exclusion, and the remaining 48.7\% are Level 1 (Supplementary).
\end{itemize}

\paragraph{Comprehensive Examination Action Space}
In the auxiliary examination phase, the model operates within a unified action space of 38 distinct categories, replicating the resource management complexity of real hospitals. This includes:
\begin{itemize}
    \item \textbf{Routine \& Biochemical:} Blood/Urine/Stool Routine, Liver/Kidney Function, Electrolytes, CRP, Glucose Metabolism (including OGTT/HbA1c), etc.
    \item \textbf{Imaging \& Functional:} CT, MRI, Ultrasound, X-ray, ECG, EEG, and Pulmonary Function Tests.
    \item \textbf{Pathology \& Specialized:} Tumor Markers, Viral Markers, Autoantibodies, Hormones, Bone Marrow Biopsy, and Endoscopy.
\end{itemize}
This extensive candidate set requires the model to precisely select necessary tests while avoiding resource waste.

\subsubsection{Workflow and Evaluation Methodology}
The evaluation pipeline starts with Station 1: Inquiry, where the model conducts multi-turn interactions with a Standardized Patient. To assess process quality, we introduce the SCAN framework, which decomposes consultation performance into four dimensions: Safety Stratification, Information Clarification, Associative Questioning, and Normative Output. We use GPT-4.1~\cite{gpt4} to verify coverage of OSCE-derived key clinical points, while excluding non-diagnostic (or easily engineered) content that is repeatedly mentioned in each consultation (e.g., age/sex restatements or templated self-introductions).


Given the elicited evidence, Station 2: Lab Testing evaluates both resource efficiency and interpretative accuracy. The model proposes a differential diagnosis and selects laboratory or imaging tests from a unified candidate pool, where candidate tests are categorized into essential and optional groups based on their contributions to clinical diagnosis and decision-making. Performance is evaluated using a weighted F1 score, in which recall for essential tests is assigned a higher weight, thereby reflecting their greater importance in the diagnostic workflow while maintaining a balanced assessment of overall precision and coverage.

The workflow concludes with Station 3: Diagnosis, where the model integrates all prior information to infer a final diagnosis. We adopt a hierarchical matching criterion based on the ICD-10 taxonomy, rewarding correct leaf-node matches while penalizing off-branch predictions.

\subsubsection{Performance Analysis}

\begin{figure}[H]
    \centering
    \includegraphics[width=1.0\linewidth]{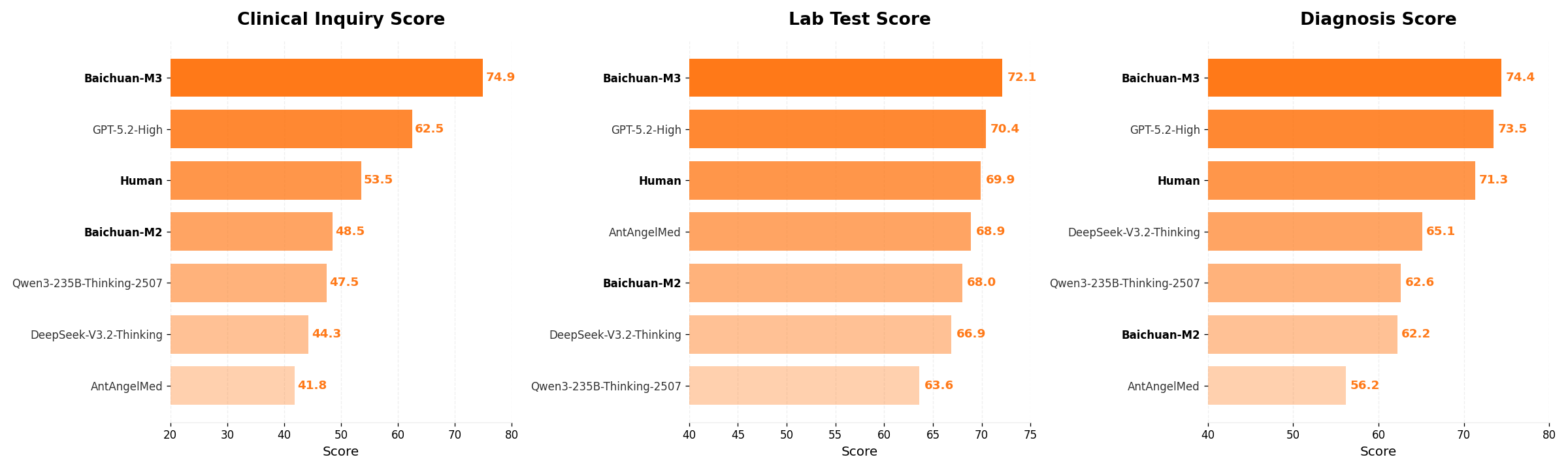}
    \caption{Overall performance comparison on ScanBench.}
    \label{fig:scan_bench_main}
\end{figure}

As illustrated in Fig.~\ref{fig:scan_bench_main}, Baichuan-M3 exhibits a comprehensive advantage, ranking first across all three stations. Most notably, in the challenging Clinical Inquiry phase, it achieves a score of 74.9, surpassing the second-best model (GPT-5.2-High) by 12.4 points and the human baseline by over 20 points. This dominance extends to laboratory testing (72.1) and final diagnosis (74.4), suggesting that Baichuan-M3 possesses robust, end-to-end medical reasoning capabilities rather than merely excelling in isolated tasks.

\begin{figure}[H]
    \centering
    \includegraphics[width=1.0\linewidth]{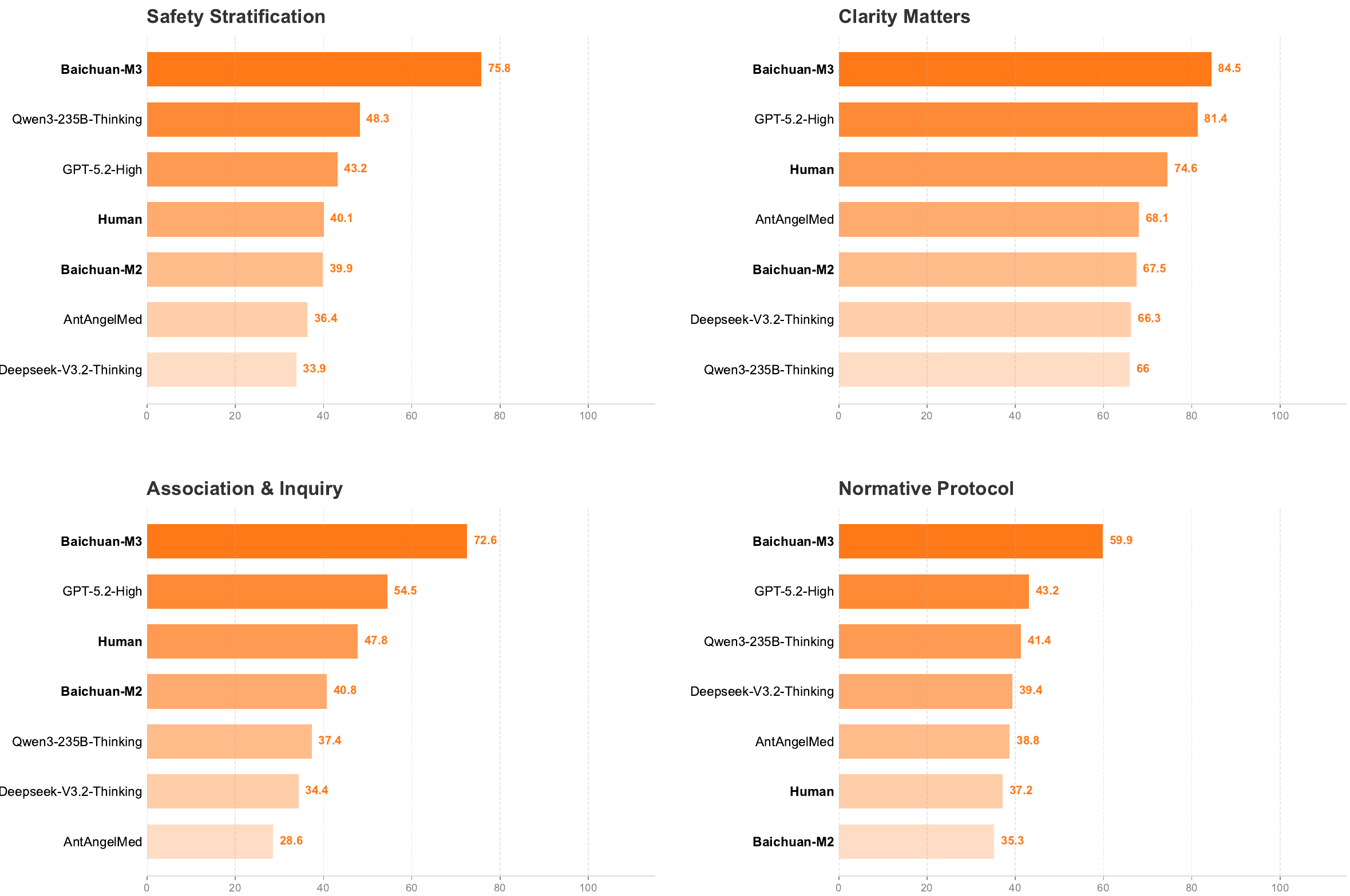}
    \caption{Detailed breakdown of Inquiry Capabilities.}
    \label{fig:scan_bench_dims}
\end{figure}

Further decomposition of the inquiry capability via the SCAN framework (Fig.~\ref{fig:scan_bench_dims}) reveals that Baichuan-M3 is the sole model to demonstrate dominant leadership across all four dimensions, consistently outperforming both SOTA LLMs and human experts.

Specifically, in Safety Stratification, Baichuan-M3 achieves a remarkable score of 75.8, creating a substantial gap over the runner-up Qwen3-235B (48.3) and nearly doubling the human benchmark (40.1). This indicates superior sensitivity to ``red flag'' symptoms and critical risks. In terms of Association \& Inquiry, the model scores 72.6, significantly surpassing GPT-5.2-High (54.5). This highlights its sophisticated grasp of differential diagnosis, enabling it to proactively uncover hidden clinical clues beyond the user's initial description. Furthermore, Baichuan-M3 excels in Clarity Matters (84.5) and Normative Protocol (59.9), ensuring that the collected information is both granular and structurally standardized. By integrating these strengths, Baichuan-M3 successfully creates a closed loop of precise inquiry and secure decision-making that exceeds human-level standardization.

\subsubsection{Dynamic Consultation Efficiency}
\begin{figure}[H]
    \centering
    \includegraphics[width=0.95\linewidth]{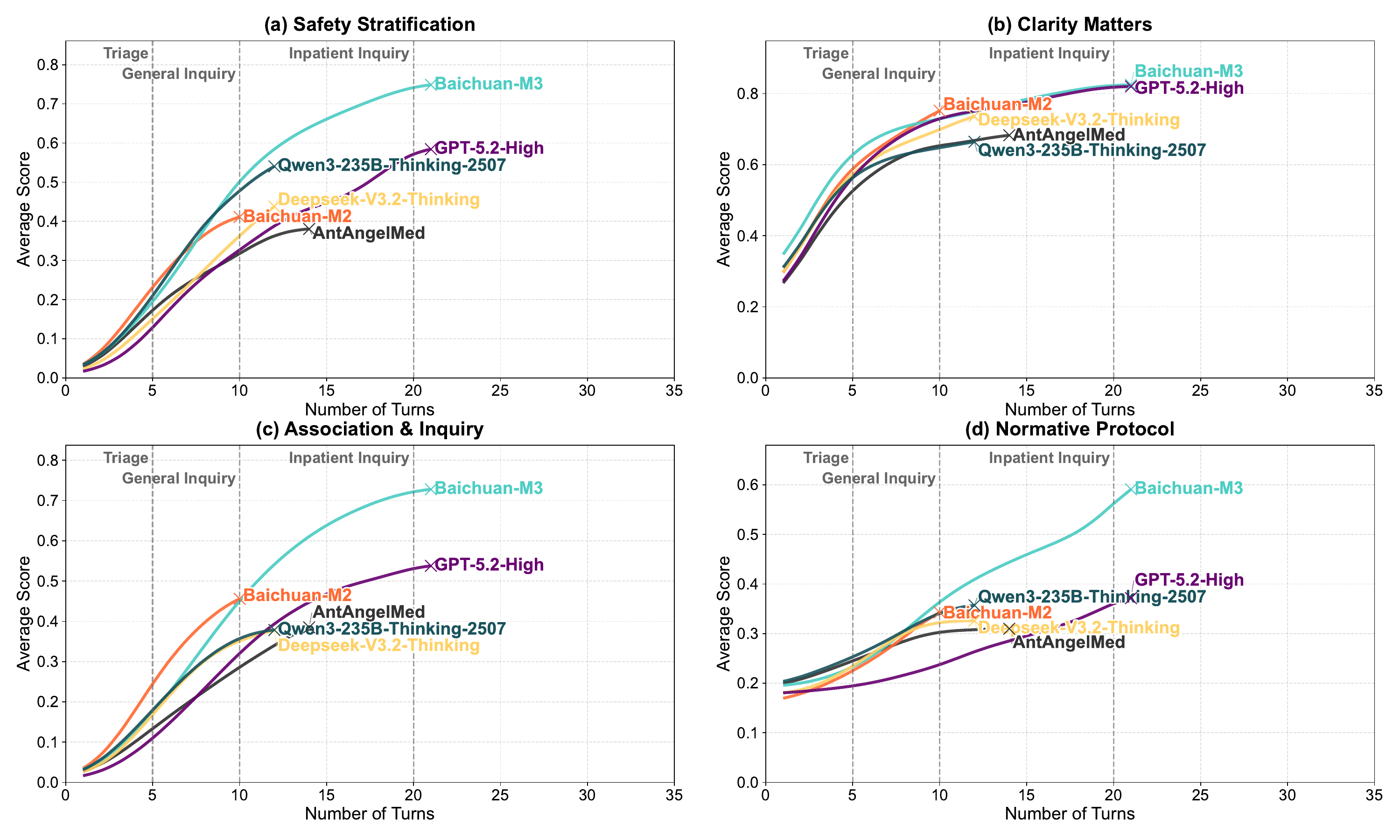}
    \caption{Evolution of model performance across dialogue turns.}
    \label{fig:scan_turn_evlo}
\end{figure}

Figure~\ref{fig:scan_turn_evlo} plots model scores against the number of dialogue turns. To reduce noise from very rare long conversations, we drop turn bins that contain fewer than 10\% of the total samples. This makes the trend across turns easier to interpret.

\paragraph{Convergence in Basics, Divergence in Reasoning}
Figure~\ref{fig:scan_turn_evlo}b shows that most models quickly reach a similar level on symptom clarification (around 0.7--0.8). The gap becomes clear in Association \& Inquiry (Fig.~\ref{fig:scan_turn_evlo}c): Baichuan-M3 keeps improving as the dialogue goes on, while generalist models (e.g., Deepseek and Qwen) fall behind — resulting in nearly a twofold advantage at longer horizons.

\paragraph{Risk Sensitivity and Contextual Adherence}
For Safety Stratification (Fig.~\ref{fig:scan_turn_evlo}a), Baichuan-M3 is more responsive to risk signals, and its score rises to about 0.7 as evidence accumulates. By contrast, GPT-5.2-High stays around 0.5 even with longer dialogues, suggesting weaker acute-risk recognition. For Normative Protocol (Fig.~\ref{fig:scan_turn_evlo}d), performance tends to improve in later turns, indicating that keeping track of context helps models follow structured clinical workflows.

In summary, while general LLMs suffice for basic information gathering, specialized training is indispensable for the complex reasoning and safety assurance required in clinical settings.

\subsection{HealthBench}
HealthBench serves as a rigorous benchmark to evaluate the clinical reasoning capabilities and safety boundaries of LLMs. By assessing performance across varying difficulty levels and dimensions, it provides a comprehensive view of a model's utility in real-world medical scenarios.

\subsubsection{HealthBench-Main}
\begin{figure}[H]
    \centering
    \includegraphics[width=0.98\linewidth]{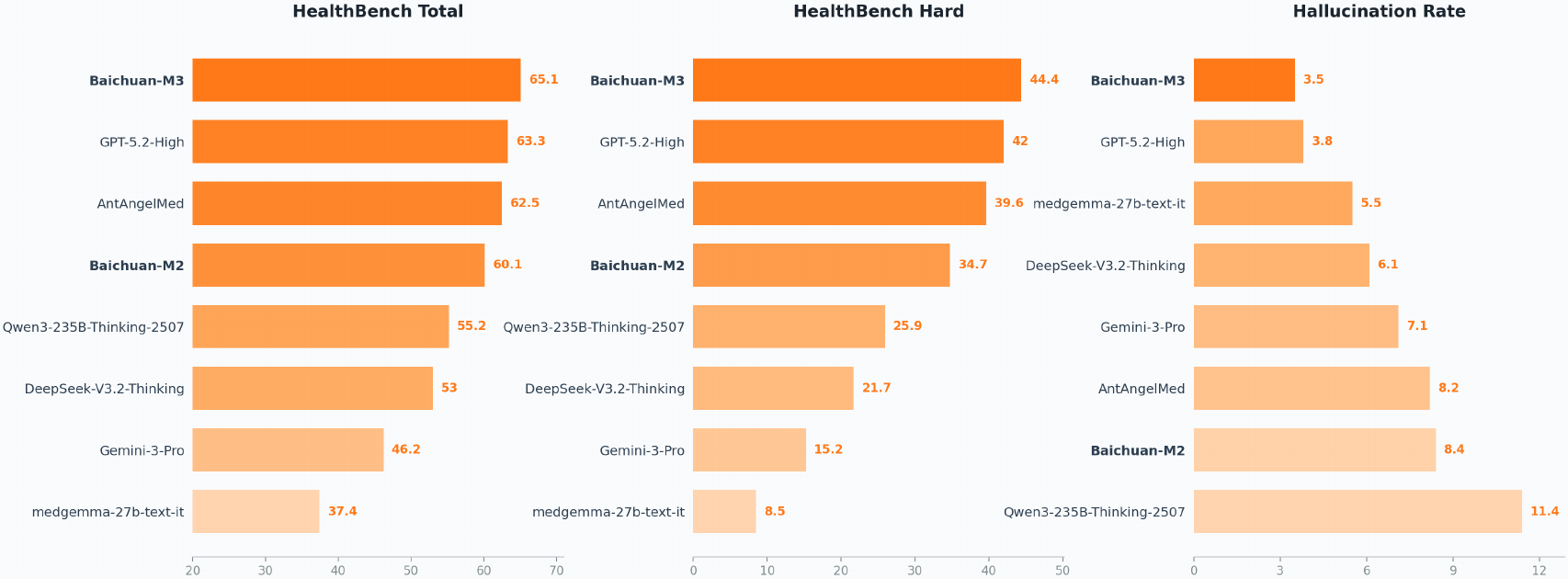}
    \caption{Overall performance comparison on HealthBench.}
    \label{fig:exp_hb}
\end{figure}

As illustrated in Fig.~\ref{fig:exp_hb}, \textbf{Baichuan-M3} establishes a new state-of-the-art (SOTA) standard across all key metrics. On the comprehensive \textit{HealthBench Total}, Baichuan-M3 achieves a score of \textbf{65.1}, surpassing the runner-up GPT-5.2-High (63.3) by a clear margin. Notably, in the challenging \textit{HealthBench Hard} subset, Baichuan-M3 further extends its lead with a score of \textbf{44.4}, significantly outperforming strong competitors such as GPT-5.2-High (42.0) and AntAngelMed (39.6). Furthermore, Baichuan-M3 demonstrates exceptional reliability by securing the lowest \textit{Hallucination Rate} of \textbf{3.5\%}, indicating a robust balance between deep clinical reasoning and factual accuracy compared to other models.

\begin{figure}[t]
    \centering
    \includegraphics[width=1.0\linewidth]{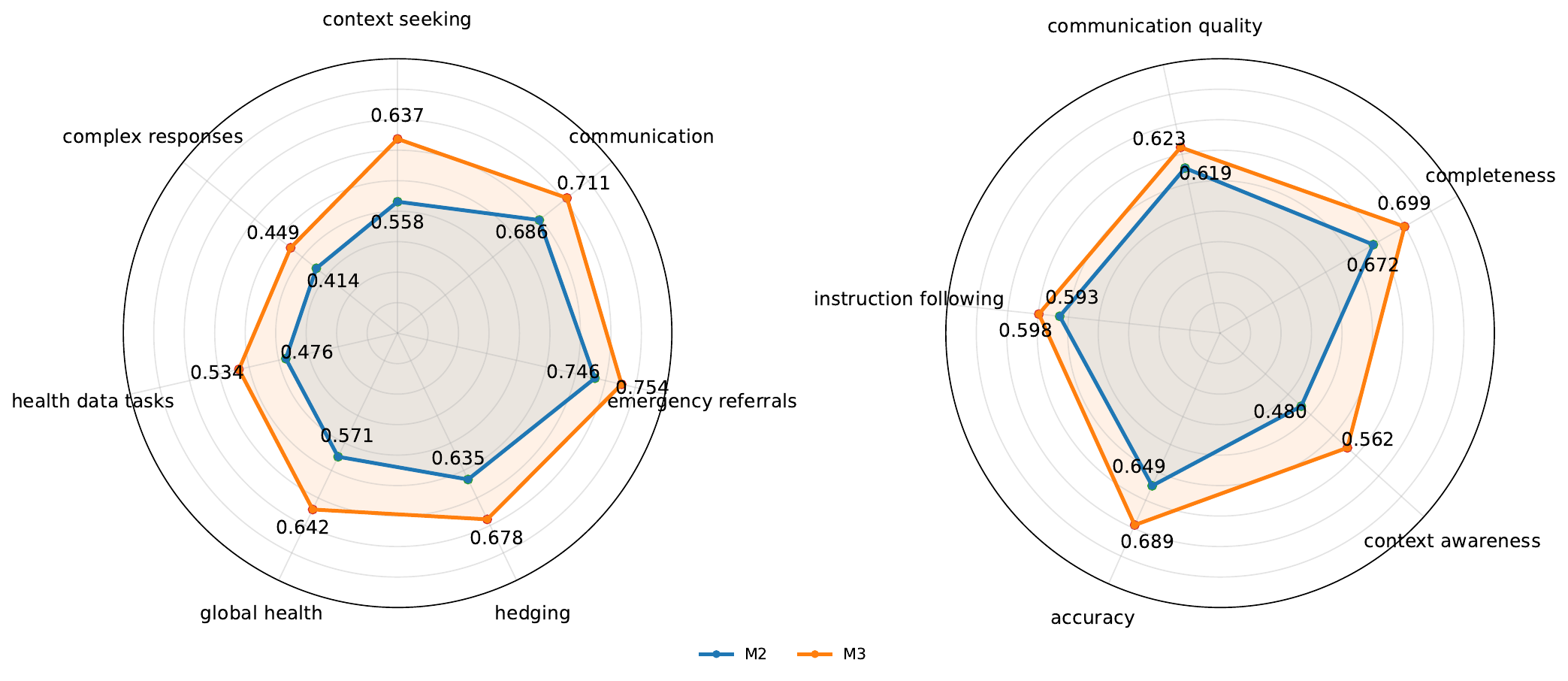}
    \caption{HealthBench fine-grained comparison: M2 vs M3 (themes \& axes).}
    \label{fig:hb_m2_m3}
\end{figure}

In terms of fine-grained capabilities compared with M2, M3 exhibits broad gains across most evaluation dimensions, resulting in a more robust and well-balanced medical service profile (see Fig.~\ref{fig:hb_m2_m3}). The improvements are especially pronounced in context seeking and context awareness. We attribute these gains to transfer from enhanced Deep Clinical Consultation training: M3 more proactively elicits missing history and risk factors while recognizing contextual constraints that shape clinical decisions. As a result, it produces more complete and reliable treatment recommendations and triage assessments. Overall, these results suggest that M3 generalizes the ``inquiry--clarification--decision'' interaction paradigm learned in consultation tasks to a broader set of healthcare scenarios.

\subsubsection{HealthBench-Hallu}

While benchmarks such as HealthBench have established standards for evaluating medical LLMs, ensuring clinical safety remains paramount for healthcare AI deployment. Medical hallucinations, which involve the generation of superficially plausible yet factually incorrect information, represent a significant concern acknowledged by regulatory authorities and the research community~\cite{ankit2023medhalt, DBLP:journals/corr/abs-2503-05777, DBLP:conf/bionlp/ManesRCBHS24}. This issue is particularly pronounced in complex medical reasoning tasks, where model outputs involve extended passages with high information density, encompassing specialized knowledge such as precise pharmaceutical dosing and rare clinical presentations. The combination of extended generation lengths and linguistic coherence creates conditions wherein fine-grained factual errors may evade detection, thereby presenting substantial clinical risks. We therefore introduce HealthBench-Hallu, a fine-grained evaluation framework designed to assess the factual integrity of model responses generated across HealthBench tasks. By decomposing these outputs into discrete atomic claims and validating them against external evidence, this metric provides a rigorous quantification of medical hallucinations.

\paragraph{Metric Design}

HealthBench-Hallu focuses on healthcare factual misconceptions embedded within the model's generated content, which primarily manifest as:

\begin{itemize}
    \item \textbf{Erroneous or misapplied medical knowledge:} Stating incorrect clinical facts or applying correct knowledge to inappropriate contexts.
    \item \textbf{Fabricated or distorted medical evidence:} Inventing non-existent references, falsifying clinical data, or fabricating causal relationships.
\end{itemize}

\paragraph{HealthBench-Hallu Evaluation}
HealthBench-Hallu reuses the Fact-Aware Verification Pipeline~(see Section~\ref{sec:fact-verifier}), but runs it in a high-precision (rather than high-throughput) configuration: we upgrade the claim extractor to GPT-5 to improve coverage of fine-grained atomic claims, and enforce real-time multi-turn search verification instead of relying on cached evidence.
Based on above evidence labels generated by this verification pipeline, we calculate the Weighted Hallucination Rate ($H$):

\begin{equation}
H = \frac{\sum_{i=1}^{N} w_i}{|\text{Total Claims}|}
\end{equation}

The weights $w_i$ are assigned based on factual risk stratification: \textbf{Refuted} (1.0) represents severe factual fallacies; \textbf{Uncertain} (0.5) represents statements with insufficient evidence or ambiguity risks; and \textbf{Supported} (0.0) denotes correct statements. This metric directly reflects the credibility boundaries of the model when addressing complex medical problems.

\paragraph{Experimental Results}

Table~\ref{tab:hallu_vs_cap} demonstrates that the integration of Fact-Aware RL enables Baichuan-M3-235B to effectively suppress hallucinations while preserving medical reasoning capabilities. The model maintains competitive task performance (HealthBench Score: 65.1) comparable to the variant without reinforcement learning (66.2), while substantially reducing both refuted response and uncertainty rates by approximately 50\%. These results indicate that the adaptive weighting strategy effectively mitigates the typical trade-off between safety constraints and model utility, avoiding the reasoning degradation commonly induced by overly conservative approaches.

\begin{table}[htbp]
  \centering
  \caption{Trade-off between hallucination suppression and capability}
    \begin{tabular}{lccc}
    \toprule
    \multicolumn{1}{c}{\textbf{Model}} & \textbf{HealthBench Score} & \textbf{Refuted Rate} & \textbf{Uncertain Rate} \\
    \midrule
    GPT-5.2-High & 63.3  & 2.37\% & 2.78\% \\
    Baichuan-M2-32B & 60.1  & 5.73\% & 5.43\% \\
    M3 - w/o Fact Aware RL & 66.2  & 4.68\% & 3.64\% \\
    \textbf{Baichuan-M3-235B} & \textbf{65.1} & \textbf{2.45\%} & \textbf{2.07\%} \\
    \bottomrule
    \end{tabular}%
  \label{tab:hallu_vs_cap}%
\end{table}%

\paragraph{Hallucination Analysis through Knowledge Probes}

\begin{figure}[h]
    \centering
    \includegraphics[width=0.98\linewidth]{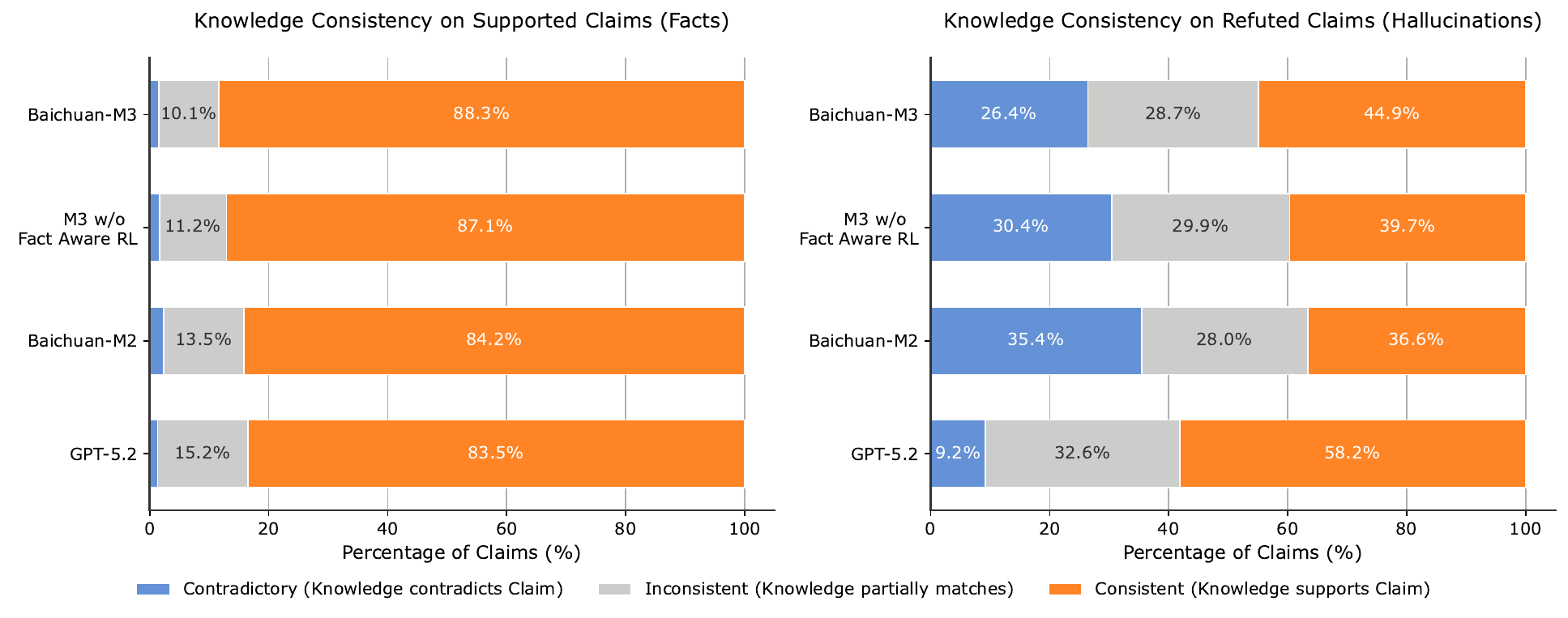}
    \caption{\textbf{Knowledge boundary alignment analysis.} We categorize the alignment between internal parameters and generated claims into three states: \textit{Consistent} (output aligns with internal cognition), \textit{Inconsistent} (partial mismatch), and \textit{Contradictory} (output opposes internal cognition). Notably, in the context of hallucinations (Refuted Claims), a \textit{Consistent} state indicates an ``honest error'' derived from incorrect internal knowledge.}
    \label{fig:Knowledge_Boundary_Alignment}
\end{figure}

To elucidate the underlying mechanism by which Fact-Aware RL reduces hallucination rates, we employ knowledge probes to analyze the alignment between the model's \textit{Internal Cognition} (parametric truthfulness) and its \textit{External Output} (generated claims), as illustrated in Fig.~\ref{fig:Knowledge_Boundary_Alignment}. This analysis allows us to decouple the sources of error by examining whether the model's output faithfully reflects its internal parameters or deviates due to generation instability.

The results reveal a distinct bifurcation in the model's alignment dynamics induced by Fact-Aware RL. For factually correct outputs (Supported Claims), the consistency between internal cognition and external output remains robust at 88.3\%, indicating that the model’s factual truths are firmly anchored in its internal parameters. More critically, for erroneous outputs (Refuted Claims), the consistency rate significantly increases to 44.9\%. This upward shift signifies a marked reduction in ``unfaithful hallucinations''—instances where the model possesses correct internal cognition but generates contradictory false information. The data suggests that the remaining hallucinations are predominantly \textit{honest errors}, where the external output faithfully reflects a misconception inherent in the model's parameters.

Ultimately, these findings imply that the primary efficacy of Fact-Aware RL lies not in injecting new knowledge, but in regulating the model's generation strategy to strictly converge within its authentic knowledge boundaries.

\section{Inference Optimization}
To facilitate the accessibility and efficient deployment of the Baichuan-M3 model in healthcare applications, two inference optimized strategies are implemented.
First, to accelerate text generation, the Gated Eagle-3 speculative decoding approach was introduced, leading to a substantial improvement in inference throughput.
Second, advanced quantization methods were applied to significantly reduce the memory requirements of the model without compromising accuracy.
These inference optimizations lower the practical barriers to deployment and support broader applications of advanced medical large language models.

\subsection{Speculative Decoding}

Speculative decoding is an inference-time acceleration technique for autoregressive generation. It uses a lightweight draft model to propose multiple candidate tokens, which are then verified in parallel by the target model. The algorithm accepts the longest verified prefix and discards the remaining candidates, allowing the target model to commit multiple tokens per verification step and thereby increasing decoding throughput.

In the Eagle-3~\cite{li2025eagle} framework, the draft model is additionally conditioned on hidden states from the target model to leverage richer semantic information. However, the pronounced capacity gap between the target and draft models can induce a representation mismatch: high-dimensional and information-dense hidden states may overwhelm the lightweight draft model, reducing its ability to effectively exploit the auxiliary signal. This often leads to lower candidate acceptance rates and consequently limits the achievable speedup.

\begin{figure}
    \centering
    \includegraphics[width=1.0\linewidth]{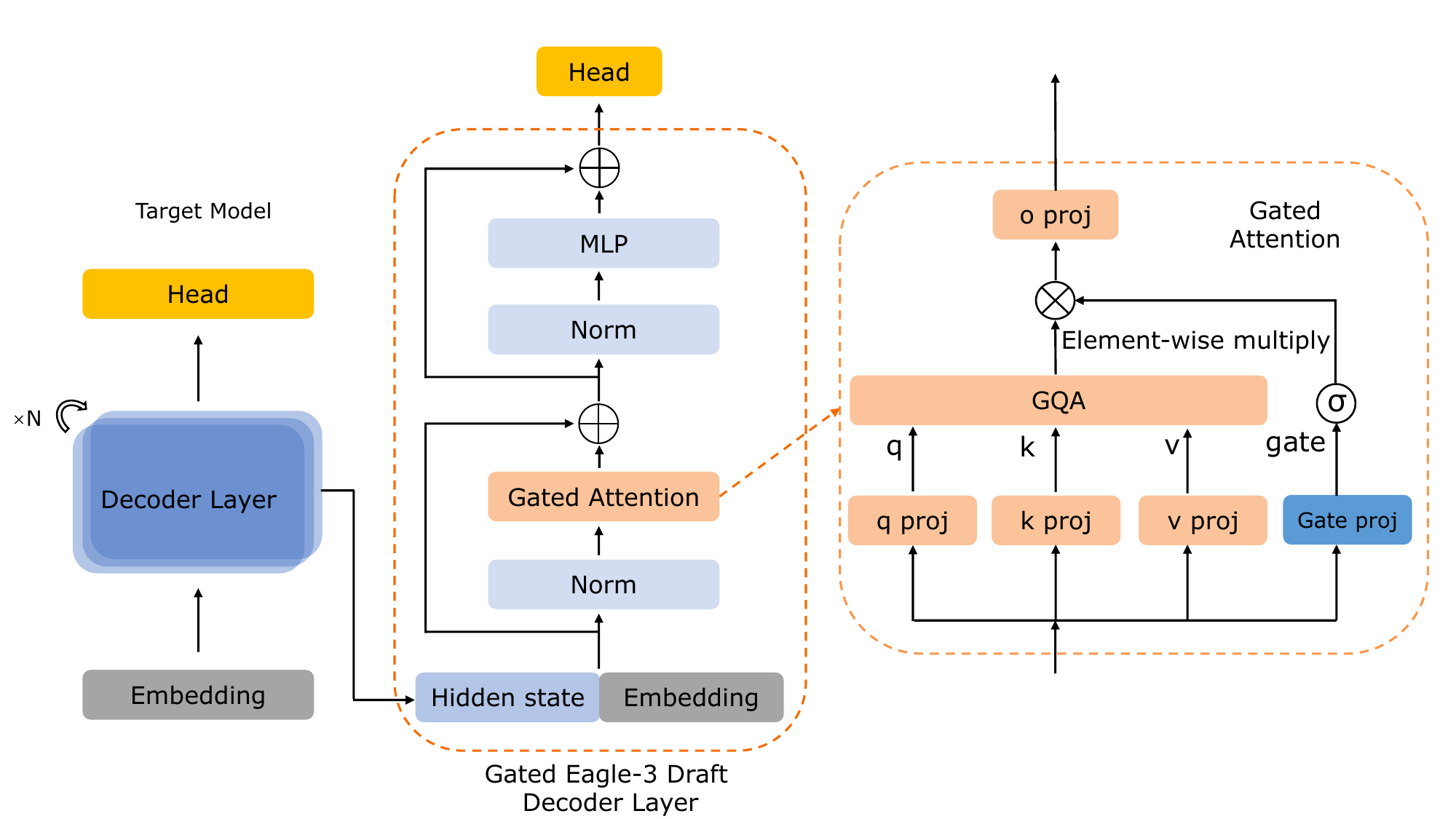}
    \caption{An illustration of Gated Eagle-3 draft model.}
    \label{fig:gated_eagle3}
\end{figure}

To mitigate this issue, we incorporate a Gated-Attention~\cite{qiu2025gated} module into the Eagle-3 draft model (called Gated Eagle-3; see Fig.~\ref{fig:gated_eagle3}), providing a dynamic and learnable mechanism to regulate the injected information. Concretely, the attention output is routed through a gating unit and modulated via element-wise multiplication with a gate vector that is dynamically generated from the current-layer input. This design enables fine-grained, dimension-wise control of information flow, emphasizing salient features while suppressing redundant or noisy components.

As detailed in Appendix~\ref{app:eagle_3}, Gated Eagle-3 achieves an average acceptance length improvement of 0.31 and improves average throughput by 12\% over the Eagle-3 base. These results indicate that selectively regulating target-model information translates more reliably into practical decoding acceleration, providing useful insights for future improvements.

\subsection{Quantization}
To reduce GPU memory consumption for Baichuan-M3 in resource-constrained settings, we apply INT4 weight quantization. However, the sparsely activated nature of Mixture-of-Experts models poses challenges for standard quantization calibration~\cite{DBLP:journals/corr/abs-2503-21135}: calibration corpora typically activate only a subset of experts, resulting in biased calibration. Consequently, frequently activated experts are quantized accurately, whereas rarely activated experts incur larger quantization errors, which can lead to unstable and unpredictable accuracy degradation at inference time.

To address this issue, we propose a self-generated calibration scheme that promotes uniform expert coverage. Specifically, we construct a multi-domain prompt set and use the BF16 model to generate high-quality responses for calibration. This approach offers two key benefits: (i) the diverse prompts activate nearly all experts, providing sufficient samples for each expert and mitigating calibration bias; and (ii) the self-generated responses encourage the quantized model to match the output distribution of the BF16 model, thereby reducing distributional discrepancies.
The training of quantization parameters was based on AutoRound~\cite{DBLP:conf/emnlp/ChengZSCHLL24} framework, and the quantization format adheres to the GPTQ~\cite{DBLP:journals/corr/abs-2210-17323} standard.

Empirically, the resulting INT4-quantized M3 model achieves near-lossless performance relative to its BF16 counterpart on mainstream benchmarks. These results validate the effectiveness of the proposed MoE-specific quantization calibration strategy and provide practical guidance for deploying large-scale sparse models.

\section{Conclusion}

We introduce \textbf{Baichuan-M3}, a medical-enhanced large language model that unifies clinical inquiry with reliable medical decision-making. By explicitly modeling the clinical workflow, Baichuan-M3 enables proactive information acquisition, coherent multi-step reasoning, and effective hallucination suppression. Through a three-stage training paradigm combining task-specific reinforcement learning and multi-teacher distillation, the model achieves strong performance on both factual and process-oriented benchmarks, including HealthBench, HealthBench-Hallu and the OSCE-style ScanBench. These results demonstrate that workflow-aligned optimization is an effective approach for advancing clinical-grade medical LLMs.

\section{Limitation and Future Work}

Baichuan-M3 is currently limited to episodic, text-based clinical scenarios and does not fully capture longitudinal disease management, multimodal clinical signals, or ultra-long-horizon reasoning across patient trajectories. While hallucination control is substantially improved, rare high-risk errors and limited explicit grounding in evidence-based sources remain open challenges. Future work will focus on extending the model toward full-pathway clinical reasoning with multimodal inputs, long-context optimization, and tighter integration of evidence retrieval, safety constraints, and environment-based reinforcement learning.

\section{Contribution}

Contributors are presented in alphabetical order according to their first names. An asterisk (*) denotes those who are no longer part of the team.

\subsection*{Core Contributors}

Chengfeng Dou, 
Fan Yang, 
Fei Li, 
Jiyuan Jia, 
Qiang Ju, 
Shuai Wang, 
Tianpeng Li, 
Xiangrong Zeng*, 
Yijie Zhou

\subsection*{Contributors}

Hongda Zhang, 
Jinyang Tai, 
Linzhuang Sun, 
Peidong Guo, 
Yichuan Mo

\subsection*{Experts and Advisors}

Xiaochuan Wang,
Hengfu Cui, 
Zhishou Zhang

\printbibliography[heading=bibintoc, title={References}]

\newpage
\appendix
\section{Appendix}
This appendix presents supplementary experimental results and detailed analyses to further validate our proposed methods. Specifically, we provide ablation studies on the SPAR algorithm and the Clip-Forward-KL objective, along with extended evaluations of the Fact-Aware RL framework and the Gated Eegle-3 speculative decoding mechanism. For comprehensive details regarding the specific prompts utilized for training and evaluation, please refer to the associated \href{https://github.com/baichuan-inc/Baichuan-M3-235B/tree/main/prompts}{GitHub} repository.

\subsection{Ablation Study of SPAR}

\begin{figure}[H]
    \centering
   \includegraphics[width=0.98\linewidth]{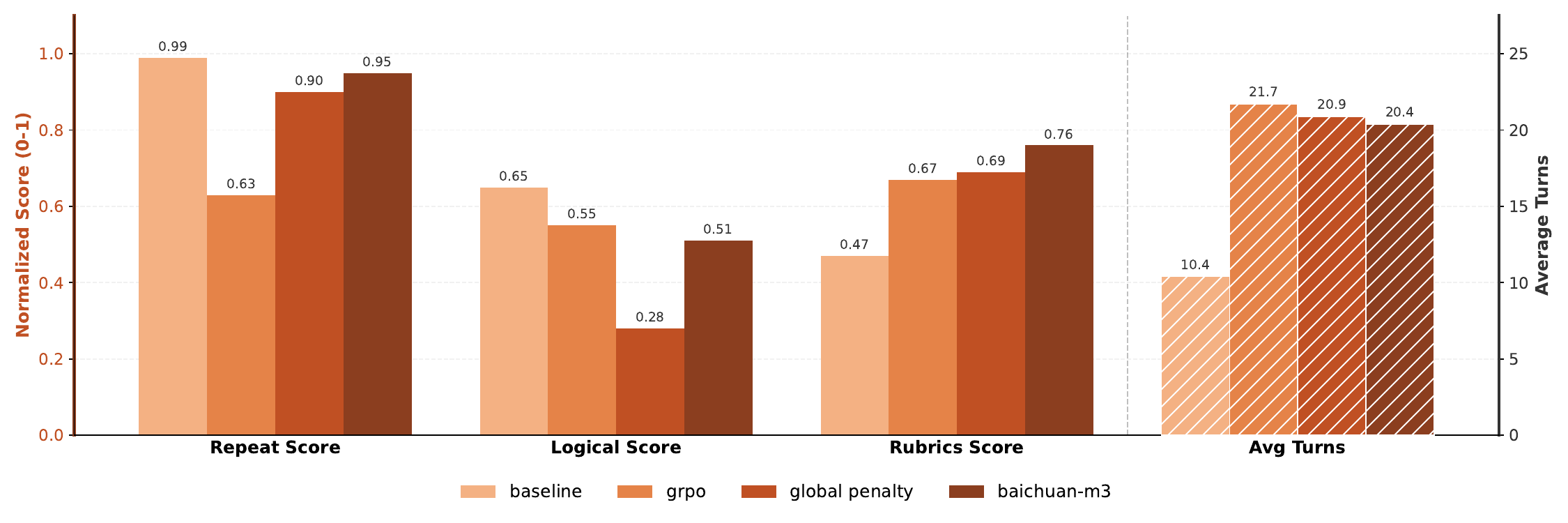}
    \caption{Performance comparison between SPAR and baseline models in multi-turn medical consultation tasks.}
    \label{fig:SPAR}
\end{figure}

To systematically evaluate the efficacy of the SPAR algorithm in multi-turn medical consultation tasks, we conducted ablation studies using the following experimental configurations:
\begin{enumerate}
\item \textbf{Backbone:} We use the Baichuan-M3 base model (before RL) as the foundational architecture.

\item \textbf{GRPO (Global Reward):} Employs the GRPO algorithm with global rewards based solely on final consultation outcomes, excluding intermediate feedback.

\item \textbf{Global Penalty:} Extends the GRPO baseline by incorporating a global repetition penalty to mitigate conversational redundancy.

\item \textbf{SPAR (Baichuan-M3):} Implements the proposed step-level reward calculation algorithm to provide granular guidance.
\end{enumerate}

\textbf{Metrics:} The evaluation framework comprises four metrics normalized to the $[0, 1]$ interval: Repeat Score (measuring non-redundancy), Logical Score, Rubrics Score, and Average Turns (Avg Turns). Specifically, the first two metrics are derived by employing GPT-5 to evaluate dialogue logs on a three-point scale $(0, 1, 2)$, with the resulting scores subsequently normalized.

\textbf{Results:} As shown in Fig.~\ref{fig:SPAR}, training with the GRPO algorithm using only global rewards yields a steady improvement in Rubrics Score, but it also increases redundant inquiries (as reflected by the decline in Repeat Score). Although adding a global repetition penalty effectively reduces redundancy, it causes a sharp drop in Logical Score, indicating severe logical fragmentation. These results suggest that coarse-grained global penalties can substantially disrupt the model's natural reasoning flow. In contrast, SPAR achieves a better balance: it markedly reduces repetition while preserving logical coherence. This dual optimization allows the model to extract a higher density of critical medical information within a limited number of consultation turns.

\subsection{Fact-Aware RL} 
\subsubsection{Evaluation of Distilled Claim Extraction Models}
\label{sec:extraction_model_evaluation}

While offline claim extraction relies on GPT-5 as the reference extractor, deploying such a large teacher model for online reinforcement learning is computationally prohibitive. To address this, we fine-tune smaller models via Supervised Fine-Tuning (SFT) to serve as efficient claim extractors during online RL. We evaluate extraction fidelity using three metrics relative to GPT-5:

\begin{itemize}
    \item \textbf{Recall:} Evaluates the coverage of the SFT model relative to the reference baseline (GPT-5).
    \item \textbf{SFT Exclusivity Rate:} The proportion of claims identified by the SFT model that are absent in the GPT-5 reference, reflecting potential over-generation or discovery of novel insights.
    \item \textbf{GPT Exclusivity Rate:} The proportion of claims identified by GPT-5 that remain uncaptured by the SFT model, indicating information loss.
\end{itemize}

\begin{table}[H]
\centering
\caption{Comparative experiments on claim extraction models across different scales.}
\label{tab:model_performance}
\begin{tabular}{lcccc}
\toprule
Model & Recall (\%) & Our Exclusive Rate (\%) & GPT Exclusive Rate (\%) & Claim Count \\
\midrule
Qwen3-8B & 30.45 & 31.02 & 69.55 & 8.50 \\
Qwen3-32B & 37.68 & 33.29 & 62.32 & 12.81 \\
SFT-A3B-30B & 69.69 & 46.89 & 30.31 & 47.19 \\
SFT-8B & 72.80 & 45.60 & \textbf{27.20} & 46.66 \\
SFT-32B & \textbf{73.00} & 47.15 & \textbf{27.01} & \textbf{46.78} \\
\bottomrule
\end{tabular}
\end{table}

As shown in Table~\ref{tab:model_performance}, SFT substantially improves extraction performance, with the 8B model achieving 72.80\% recall compared to 30.45\% for the untuned baseline. While the 32B variant provides marginal performance gains, the significantly higher deployment cost is not justified for the online RL pipeline. We adopt the SFT-8B model as the claim extractor, balancing extraction fidelity with practical deployment constraints.

\subsubsection{Ablation Study on Reward Components}
\label{sec:ablation_reward}

We investigate the impact of different reward shaping strategies by comparing our method against two control groups starting from an intermediate SFT checkpoint: a \textit{w/o Fact Aware RL} model optimizing only task rewards, and a \textit{Baseline} utilizing static hallucination penalties in Eq.~\eqref{eq:naive-objective}.

\begin{figure}[H]
    \centering
    \includegraphics[width=0.98\linewidth]{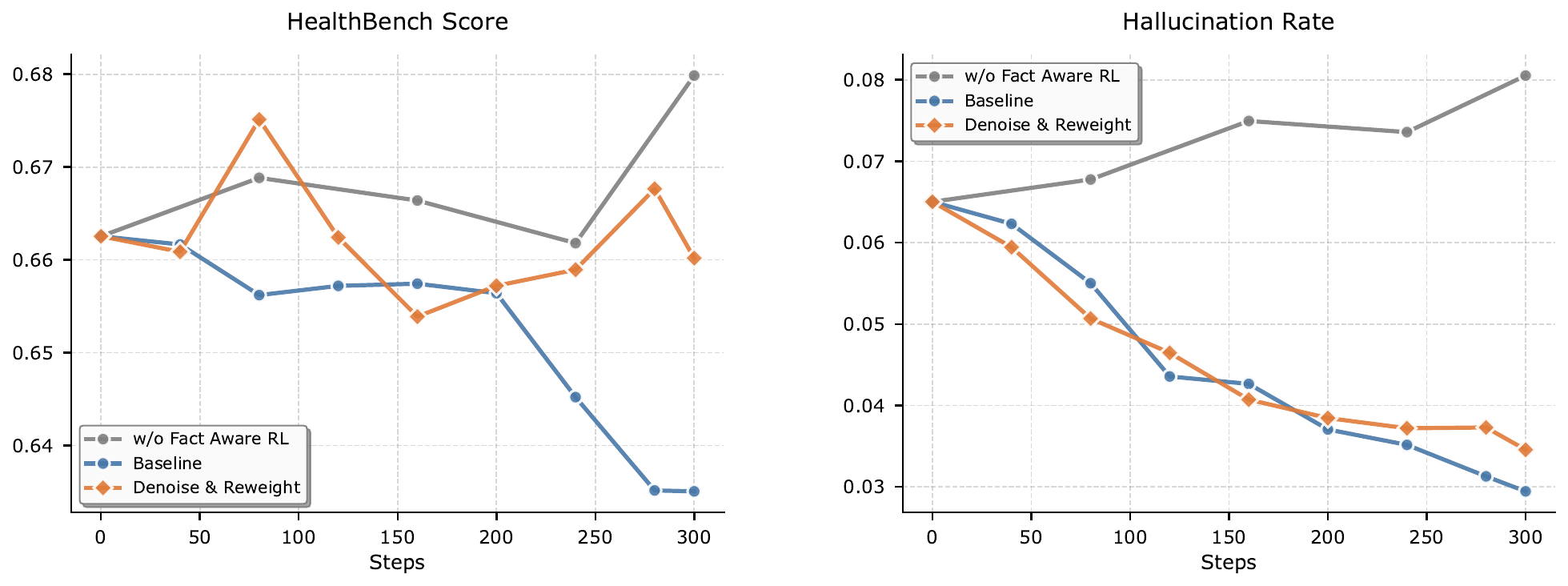}
    \caption{Training dynamics of different optimization strategies. The plots contrast the impact of each objective on medical reasoning capability (Left) and factual reliability (Right).}
    \label{fig:training_dynamics}
\end{figure}

Figure \ref{fig:training_dynamics} highlights distinct behavioral patterns driven by these objectives. The \textit{w/o Fact Aware RL} model (Grey) achieves the highest HealthBench score ($\sim$0.68) but suffers from significant hallucination drift (increasing to 0.08). This observation suggests that an unconstrained optimization objective tends to bias the model toward generating expansive content, which may inadvertently compromise factual stability. The \textit{Baseline} (Blue) corrects this drift but overcompensates; its aggressive suppression of hallucinations results in a severe degradation of reasoning capability, characteristic of penalty-induced conservatism.

Our \textit{Denoise \& Reweight} strategy in Eq.~\eqref{eq:full-fact-aware-rl} effectively decouples safety alignment from capability loss. It achieves a hallucination reduction rate comparable to the Baseline (dropping to $\sim$0.035) while maintaining reasoning scores ($\sim$0.665) close to the unconstrained model's starting point. This stability verifies the efficacy of our dual-protection design, where marginal noise filtering and competence-based gating work in tandem. Consequently, the approach successfully guides the model toward factual correctness while preserving its core medical utility.

\subsection{Ablation Study on Clip-Forward-KL for Offline Expert Fusion}
\label{sec:ablation_clip_fkl}

We analyze the role of Clip-Forward-KL in offline expert fusion. 
The student model is initialized from a \textit{medical inquiry expert} and incorporates a \textit{healthcare expert} through offline distillation. 
Both inquiry and healthcare data are distilled using either Clip-Forward-KL or standard Forward-KL, under identical initialization, datasets, and training configurations.

Table~\ref{tab:clip_ablation} reports results on ScanBench, HealthBench, and HealthBench-Hard. 
ScanBench primarily reflects the preservation of medical inquiry capability, 
while HealthBench and HealthBench-Hard assess the effectiveness of integrating broader healthcare expertise.

\begin{table}[H]
\centering
\caption{Ablation on Clip-Forward-KL for offline expert fusion. 
Clip-Forward-KL improves HealthBench and HealthBench-Hard while maintaining comparable ScanBench performance under identical training configurations.}
\label{tab:clip_ablation}
\begin{tabular}{lccc}
\toprule
Method & ScanBench & HealthBench & HealthBench-Hard \\
\midrule
Forward-KL & 73.7 & 58.6 & 33.2 \\
Clip-Forward-KL & 73.5 & \textbf{61.1} & \textbf{38.5} \\
\bottomrule
\end{tabular}
\end{table}

Clip-Forward-KL preserves inquiry performance while substantially improving healthcare benchmarks, indicating more effective fusion of new domain expertise. 
This suggests that standard Forward-KL tends to over-amplify probabilities on sparse offline samples, which interferes with the integration of new healthcare capabilities rather than the retention of existing inquiry ability. 
By enforcing a one-sided lower-bound constraint on teacher-supported actions, Clip-Forward-KL enables a more conservative fusion and yields an initialization for on-policy optimization.

\subsection{Ablation Study on Gated Eagle-3 Speculative Decoding.}

To ensure a rigorous and fair comparison, we evaluated the original Eagle-3 draft model against our proposed Gated Eagle-3 using an identical dataset of 35,000 training samples and the same training protocol. Both models were assessed under uniform inference configurations across a suite of representative benchmarks, including GSM8K~\cite{DBLP:journals/corr/abs-2110-14168}, HumanEval~\cite{DBLP:journals/corr/abs-2107-03374}, MT-Bench~\cite{DBLP:conf/nips/ZhengC00WZL0LXZ23}, and HealthBench. All experiments were conducted on NVIDIA H20 GPUs, with model training performed using the SpecForge~\cite{specforge2025} framework and inference executed via SGLang~\cite{zheng2024sglang}. As shown in Table~\ref{tab:eagle_comparison}, Gated Eagle-3 achieves an average acceptance length improvement of 0.31 over the Eagle-3 base. 
Moreover, throughput comparisons conducted under a parallelism of 8 are presented in Table~\ref{app:eagle_throughput_comparison}.

\label{app:eagle_3}
\begin{table}[H]
\centering
\caption{Comparison of average acceptance length between Eagle-3 Base and Gated Eagle-3.}
\label{tab:eagle_comparison}
\begin{tabular}{lcccc}
\toprule
Method & GSM8K & HumanEval & MT-Bench & HealthBench  \\
\midrule
Eagle-3 Base & 3.59 & 3.58 & 2.97 & 2.41 \\
Gated Eagle-3  & \textbf{3.84} & \textbf{4.03} & \textbf{3.23} & \textbf{2.70}  \\
\midrule
$\Delta$ (Gain) & +0.25 & +0.45 & +0.26 & +0.29  \\
\bottomrule
\end{tabular}
\end{table}

\begin{table}[H]
\centering
\caption{Comparison of throughput~(tokens/s) between Eagle-3 Base and Gated Eagle-3.}
\label{app:eagle_throughput_comparison}
\begin{tabular}{lcccc}
\toprule
Method & GSM8K & HumanEval & MT-Bench & HealthBench  \\
\midrule
Eagle-3 Base & 376.98 & 576.19 & 451.39 & 356.94  \\
Gated Eagle-3 & \textbf{431.26} & \textbf{646.17} & \textbf{490.12} & \textbf{400.52}  \\
\midrule
$\Delta$ (Gain) & +14.40\% & +12.15\% & +8.58\% & +12.21\%  \\
\bottomrule
\end{tabular}
\end{table}


\addappheadtotoc

\end{document}